\RequirePackage{fix-cm}
\documentclass{article}

\PassOptionsToPackage{table,dvipsnames}{xcolor}
\usepackage{xcolor}

\usepackage[preprint]{corl_2026}

\hypersetup{
    pdftitle={DiffMorphGrasp: A Diffusion Framework Conditioned on Hand Morphology for Cross-Embodiment Grasp Generation},
    pdfauthor={Zhiyuan Wu, Xiangyu Zhang, Zhuo Chen, Jiankang Deng, Rolandos Alexandros Potamias, Shan Luo}
}

\usepackage{bbding}
\usepackage{graphicx}
\usepackage{subcaption}
\usepackage{booktabs}
\usepackage{multirow}
\usepackage{gensymb}
\usepackage{textcomp}
\usepackage{arydshln}

\usepackage{amsmath}
\usepackage{amssymb}
\usepackage{mathtools}
\usepackage{amsthm}
\usepackage{pifont}
\usepackage{wrapfig}

\usepackage{hyperref}
\usepackage[capitalize,noabbrev]{cleveref}

\theoremstyle{plain}

\theoremstyle{definition}

\theoremstyle{remark}

\newcommand{\zhiyuanrevision}[1]{\textcolor{black}{#1}}

\title{DiffMorphGrasp: A Diffusion Framework Conditioned on Hand Morphology for Cross-Embodiment Grasp Generation}

\author{
  Zhiyuan Wu \\
  King's College London \\
  \texttt{zhiyuan.1.wu@kcl.ac.uk}
  \And
  Xiangyu Zhang \\
  King's College London \\
  \texttt{xiangyu.zhang@kcl.ac.uk}
  \And
  Zhuo Chen \\
  King's College London \\
  \texttt{zhuo.7.chen@kcl.ac.uk}
  \AND
  Jiankang Deng \\
  Imperial College London \\
  \texttt{j.deng16@imperial.ac.uk}
  \And
  Rolandos Alexandros Potamias\textsuperscript{\Envelope} \\
  Imperial College London \\
  \texttt{r.potamias@imperial.ac.uk}
  \And
  Shan Luo\textsuperscript{\Envelope} \\
  King's College London \\
  \texttt{shan.luo@kcl.ac.uk} \\
}

\begin{document}
\maketitle

\begin{figure*}[h!]
    \centering
    \vskip -2.4em
    \includegraphics[width=\linewidth]{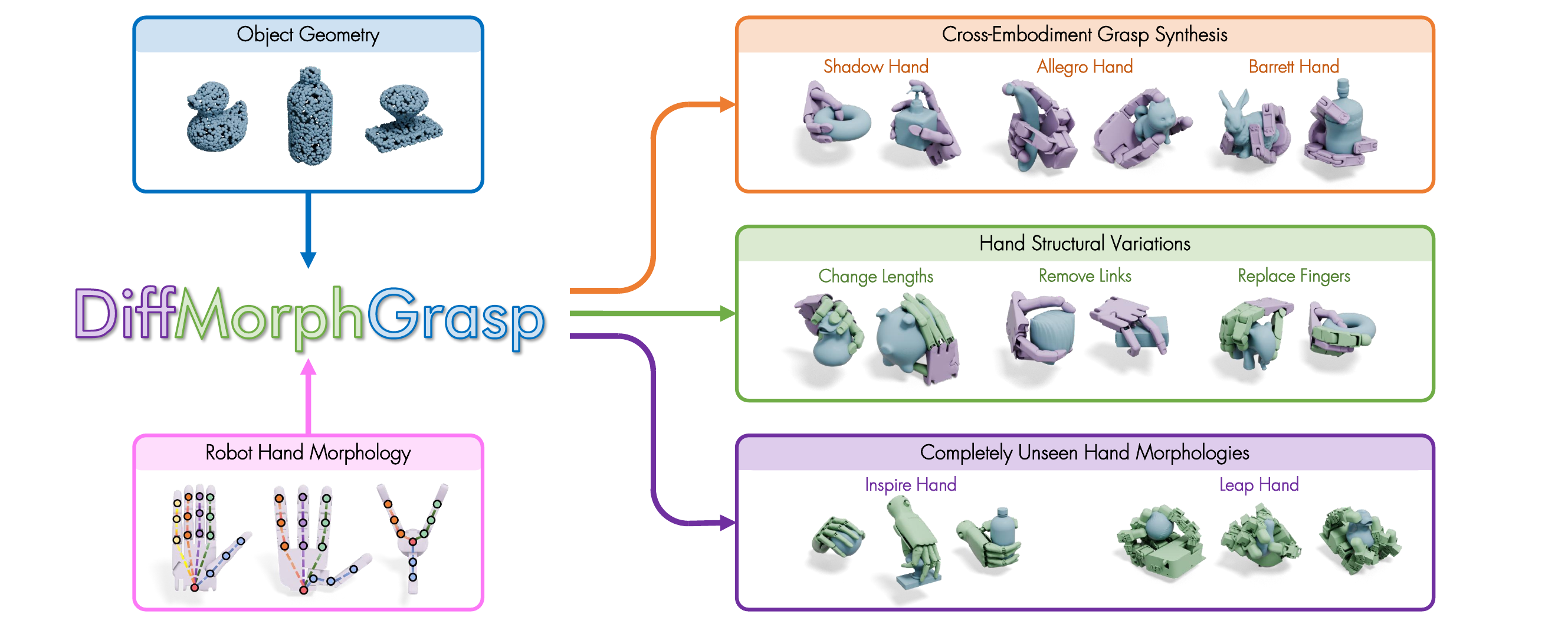}
    \vskip -0.4em
    \caption{ 
        Given object geometry and robot hand morphology, DiffMorphGrasp generates cross-embodiment grasps, generalizing from training hand embodiments (purple hand parts) to novel hand-structure variations and unseen hand morphologies (green hand parts).
    }
    \label{fig:teaser}
    \vskip -0.8em
\end{figure*}


\begin{abstract}
As robots are deployed across diverse environments, manipulation tasks require robot hands with different morphologies to satisfy varying task requirements, hardware capabilities, and cost constraints.
However, transferring grasping policies learned on one robot hand to another with a different morphology remains a fundamental challenge due to variations in kinematic structures and degrees of freedom across embodiments. 
To address these limitations, we propose \textbf{DiffMorphGrasp}, the first diffusion-based framework that incorporates hand morphology information into the generative process for cross-embodiment dexterous grasping.
The proposed approach maps grasps from diverse robotic hands into a reference hand pose, providing a common space for learning. Grasp generation is then continuously conditioned on structured representations of hand kinematics, encoded as graphs derived from hand configurations, together with object geometry. In addition, we introduce a loss function that computes joint-level errors over the hand kinematic structure to constrain the generated grasps during training. 
To evaluate DiffMorphGrasp, we construct a new benchmark for assessing grasp generation performance across robot hands with different morphologies. Extensive experiments demonstrate that DiffMorphGrasp achieves strong zero-shot generalization to unseen hand morphologies, enabling scalable cross-embodiment grasp deployment. \textcolor{blue}{Project page: \url{https://georgewuzy.github.io/DiffMorphGrasp-website/}}
\end{abstract}

\keywords{Robot Grasping, Cross-Embodiment, Grasp Generation} 


\section{Introduction}
Robotic manipulation systems are being introduced into a wide range of application domains, where grasping requirements differ across environments and tasks. For example, warehouse logistics often favors parallel grippers or grippers of low degrees of freedom for robust object picking and fast operation, while household service robots require multi-fingered dexterous hands to manipulate articulated tools and deformable objects in cluttered environments \cite{li2023gendexgrasp}. As a result, robot hands exhibit substantial morphological diversity in the number of fingers, joint configurations, and kinematic structures, ranging from human-like five-fingered Shadow hands to non-anthropomorphic three-fingered Barrett hands \cite{wu2025cedex}. This hardware heterogeneity poses a major challenge for transferring grasping capabilities across robot hand embodiments, even when the hands are designed for the same manipulation tasks \cite{wei2024drograsp}.

Recent advances in dexterous grasp generation have led to large-scale datasets and learning-based frameworks that improve grasp diversity and stability for robotic hands. Existing works have explored simulation-based grasp dataset generation \cite{wang2023dexgraspnet}, proposal-based grasp synthesis \cite{xu2023unidexgrasp,wan2023unidexgrasp++},  and physics-aware grasp generation frameworks \cite{zhong2025dexgraspanything, wu2025cedex}. Other approaches have also investigated grasp synthesis for robotic hands with different structures through force-closure objectives and differentiable contact modeling \cite{liu2021dfc}. 

Despite these advances, most existing methods represent robot hands using embodiment-specific joint parameterizations, requiring separate grasp representations and policies for different hands. Such representations do not explicitly model the underlying kinematic relationships between joints and fingers, making it difficult to transfer grasping knowledge across embodiments with different kinematic structures and degrees of freedom. Consequently, grasp generation models trained on one robotic hand often fail to generalize to unseen hand morphologies. 
Dexterous robotic hands exhibit structured kinematic organization that can be modeled as graphs, where joints form nodes and kinematic chains define edges. 
\zhiyuanrevision{Recent works \cite{patel2025getzero, zhang2025machagrasp} have shown the potential of graph encoders for the hand morphology through joint regression from given robot states. However, diverse grasp synthesis requires moving beyond direct regression toward generative modeling, where hand kinematics shape the grasp distribution rather than only predicting a single joint configuration.}
To this end, we propose \textbf{DiffMorphGrasp}, the first diffusion-based framework that incorporates hand morphology information into the generative process for cross-embodiment dexterous grasping, enabling zero-shot generalization to unseen robot hand structures. By continuously conditioning the generative process with learned hand kinematic morphology, the model is able to adapt grasp distributions to previously unseen hand structures, while retaining the expressive capability of diffusion models to generate diverse and stable grasps. 

As shown in Fig.~\ref{fig:teaser}, given a target dexterous hand and an object point cloud, our goal is to generate stable and diverse grasp poses that generalize across embodiments. To enable cross-embodiment learning, grasp poses of different hand structures are first mapped into a reference five-finger kinematic tree. Grasp synthesis is then performed in this space using a diffusion model conditioned on both hand structure and object geometry. Specifically, hand kinematics are extracted from URDF descriptions and encoded as graph-based morphology representations, which continuously condition the iterative denoising process through actual kinematic structure rather than discrete hand classes. This enables the model to generate object-appropriate and kinematically feasible grasps for novel hand morphologies beyond the training distribution.
Importantly, the 24-DoF mapping is only a reference for the hand pose representation rather than a humanoid-shape constraint, allowing DiffMorphGrasp to learn diverse hand kinematics through continuous conditioning and generalize to both humanoid and non-humanoid robot hands.
In addition, a loss function is introduced to adaptively scale joint errors based on their hierarchical positions in the kinematic tree using a non-linear geometric mean formulation, balancing proximal and distal gradients while enforcing kinematic constraints across diverse topologies during training. Extensive experiments demonstrate that our approach achieves state-of-the-art performance on multiple benchmarks and exhibits strong generalization to previously unseen hand structures.

Our contributions can be summarized as follows:
1) We propose \textbf{DiffMorphGrasp}, the first diffusion-based framework conditioned on hand morphology for cross-embodiment dexterous grasping, which enables zero-shot generalization to unseen robot hand structures. 
2) We introduce a novel generalization evaluation setting that covers topological, geometrical, and hybrid embodiment variations, and completely unseen hands, providing a comprehensive testbed for cross-embodiment robustness.
3) DiffMorphGrasp consistently outperforms existing methods on three established benchmarks and achieves state-of-the-art (SoTA) performance.

\section{Related Works}

\noindent \textbf{Dexterous Grasp Generation.} 
Dexterous grasping acts as an essential element for various complex manipulation tasks. Early approaches \cite{ferrari1992planning, ponce1993characterizing, miller2004graspit, prattichizzo2012manipulability, rosales2012synthesis} employ analytical methods based on contact mechanics, suffering from prohibitive computational costs due to high-dimensional optimization. Previous data-driven methods can be divided into regression-based methods \cite{liu2020ddg, xu2024dexterous} that directly predict grasp poses, contact-based methods \cite{jiang2021grasptta, xu2023unidexgrasp} that model contact point distributions, and demonstration-based methods \cite{taheri2020grab, liu2024realdex} that retarget human motions. However, these methods are inherently \textit{hand-specific}, training exclusively for a particular hand configuration. When confronted with novel hand structures, they either require substantial model re-design and retraining or completely fail to generalize. 
To address the above issue, \textit{cross-embodiment} dexterous grasp generation has emerged as a research direction, aiming to adaptively generate grasps for diverse robotic hands with varying structures utilizing a single unified model. 
Optimization-based methods \cite{liu2021dfc, li2023gendexgrasp} incorporate physical constraints, including force closure, to generate grasps for diverse hand configurations. Nevertheless, the iterative nature of the optimization introduces significant computational costs, prohibiting their real-time applications. 
Representation learning-based methods such as GeoMatch series \cite{attarian2023geomatch, wei2024geomatch++} and DRO-Grasp \cite{wei2024drograsp} learn data-driven representations by predicting intermediate contact point clouds followed by optimization-based joint angle computation. However, their reliance on regression-based prediction from point clouds limits grasp diversity, and their generalization is constrained to hand structures within the training distribution, failing when confronted with significantly different kinematic structures \cite{wei2024drograsp}.

Recently, diffusion models have emerged as a promising paradigm for dexterous grasp generation, driven by their stable training objectives that enable the effective modeling of complex, multimodal grasp distributions. Pioneering works such as Scene-Diffuser \cite{huang2023scenediffuser} and UGG \cite{lu2024ugg} first introduced diffusion models to enhance grasp diversity and object generalization.  DexGrasp-Diffusion \cite{zhang2024dexgraspdiffusion} extends this by incorporating hand class tokens to adapt to different hand types within a fixed set. More recently, DexGrasp-Anything \cite{zhong2025dexgraspanything} integrates physics-aware constraints into both training and inference for improved grasp stability. 

\noindent \textbf{Morphology Representation.} 
Morphology representation is essential for generalizable robot control across diverse embodiments. Early works \cite{wang2018nervenet, pathak2019learning} employ Graph Neural Networks (GNNs) to encode kinematic structures as graphs with limbs as nodes and joints as edges. Follow-up studies \cite{kurin2020my, guptametamorph} demonstrate that Transformers with full attention outperform GNN-based methods and achieve zero-shot generalization by treating morphology as a learned modality rather than hard-coded graph constraints. More recent works \cite{patel2025getzero, zhang2025machagrasp} introduce embodiment-aware transformers that regress joint configurations from given robot states.

\section{Methodology}

\begin{figure*}[t!]
    \centering
    \includegraphics[width=\linewidth]{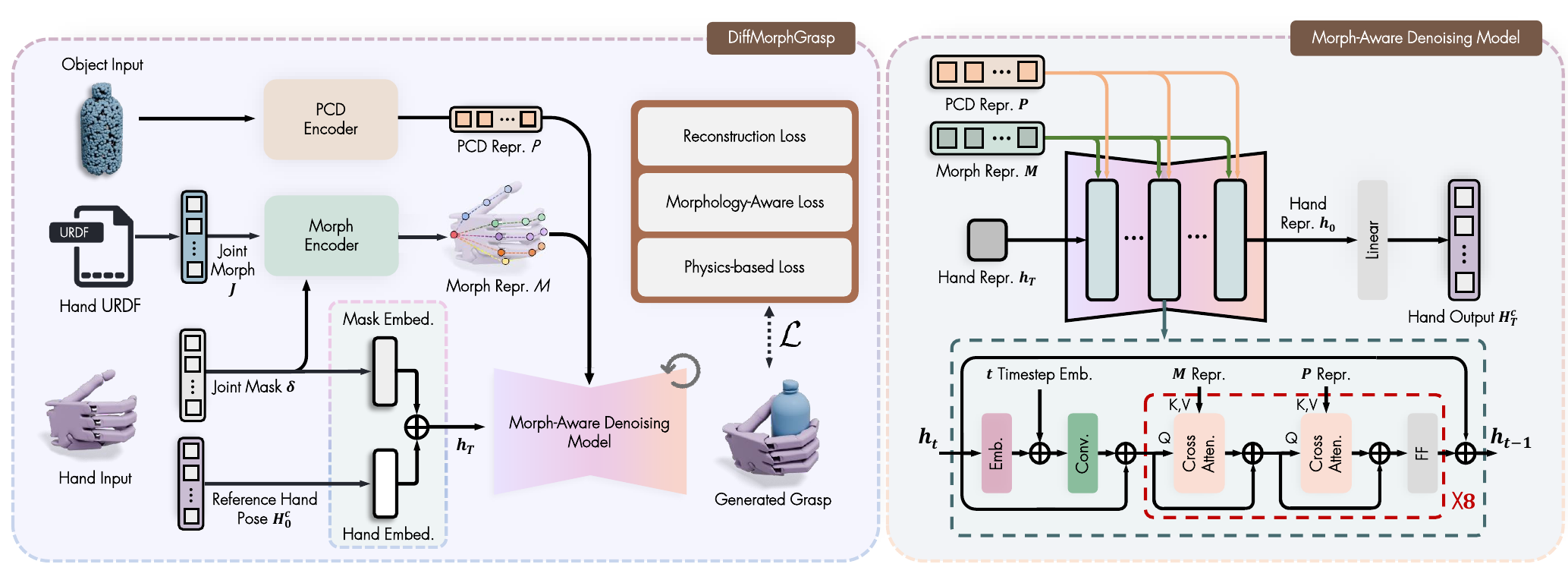}
    \vskip -0.4em
    \caption{ 
    \textbf{(Left)} The overview of our proposed DiffMorphGrasp for cross-embodiment dexterous grasp generation. Given an object point cloud and a target hand morphology extracted from its URDF specification (mapped to a pre-defined reference hand pose), we employ a morphology encoder to extract morphology representations from the hand's joint structure. The hand pose (noised via a diffusion scheduler in training) is embedded through a linear layer, and concatenated with its active joint mask embedding to obtain the hand pose representation. This representation is then processed through a morphological denoising model, where the iterative process is conditioned on both the morphology representation and the point cloud representation extracted via a Point Transformer~\cite{zhao2021pointtransformer}. The entire framework is trained based on a morphological loss function. \textbf{(Right)} The structure of our morphological denoising model, which is conditioned on the encoded morphology and the point cloud representations via cross-attention. 
    }
    \label{fig:pipeline}
    \vskip -0.4em
\end{figure*}

Fig.~\ref{fig:pipeline} illustrates the pipeline of our proposed DiffMorphGrasp, which generates stable and diverse grasp poses given a target dexterous hand URDF file and an object point cloud. The following subsections detail our problem formulation, reference hand mapping, the designs of our morphology encoder, morphological denoising model, and morphological loss function. 

\newpage

\noindent \textbf{3.1. Problem Formulation} \label{sec:problem_formulation}

Given an object point cloud $\boldsymbol{O} \subset \mathbb{R}^{3}$ and the morphology $\mathcal{M}$ of a target hand embodiment, extracted from its URDF specification, we aim to sample a batch of dexterous grasp poses $\boldsymbol{H}$ from a conditional distribution $P(\boldsymbol{H}|\boldsymbol{O}, \mathcal{M})$. 
Each pose is parametrized as $(\boldsymbol{t}, \boldsymbol{R}, \boldsymbol{\theta}) \in \mathbb{R}^{9+N}$, consisting of a global translation $\boldsymbol{t} \in \mathbb{R}^3$, a global rotation $\boldsymbol{R} \in \mathbb{R}^6$, and joint angles $\boldsymbol{\theta} \in \mathbb{R}^N$, where $N$ denotes the number of joint degrees of freedom (DoFs) (\textit{e.g.}, 24 for Shadow hand). The conditional distribution $P(\boldsymbol{H}|\boldsymbol{O}, \mathcal{M})$ is modeled using a diffusion model $\epsilon_{\phi}(\boldsymbol{H}_t, \boldsymbol{O}, \mathcal{M}, t)$, which iteratively transforms an isotropic Gaussian distribution $\mathcal{N}(0, I)$ into the desired data distribution:
\begin{equation}
    P(\boldsymbol{H}_0|\boldsymbol{O}, \mathcal{M}) = P(\boldsymbol{H}_T) \prod_{t=1}^{T} P(\boldsymbol{H}_{t-1}|\boldsymbol{H}_t, \boldsymbol{O}, \mathcal{M}), 
\end{equation}
where
\begin{equation}
P(\boldsymbol{H}_{t-1}|\boldsymbol{H}_t, \boldsymbol{O}, \mathcal{M}) = \mathcal{N}(\boldsymbol{H}_{t-1}; \mu_{\phi}, \Sigma_{\phi}),
\end{equation}
with $\mu_{\phi} \in \mathbb{R}^{9+N}$ and $\Sigma_{\phi} \in \mathbb{R}^{(9+N) \times (9+N)}$ predicted by the denoising diffusion model $\epsilon_{\phi}$ conditioned on $\boldsymbol{H}_t$, $\boldsymbol{O}$, $\mathcal{M}$, and $t$.



\noindent \textbf{3.2. Reference Hand Mapping} \label{sec:unified_hand_mapping}

\begin{figure*}[t!]
    \centering
    \includegraphics[width=0.9\linewidth]{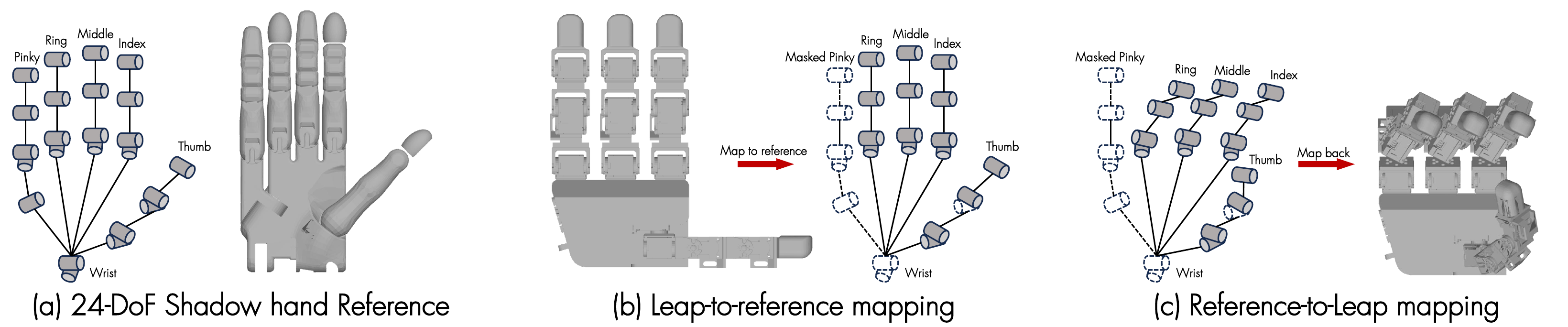}
    \vskip -0.4em
    \caption{
    Reference hand mapping example.
    (a) The 24-DoF Shadow hand serves as a fixed reference.
    (b) The Leap hand is mapped to the reference through chain/joint correspondence, with missing joints masked.
    (c) Generated active reference joints are mapped back to the Leap hand.
    }
    \label{fig:reference_mapping}
    \vskip -0.4em
\end{figure*}

To accommodate different hand morphologies within a fixed-dimensional generative model, we follow \cite{zhang2024dexgraspdiffusion} and use the 24-DoF Shadow hand as a fixed reference. This differs from \cite{wei2026one}, which constructs a canonical URDF and parameterization, while our framework explicitly encodes morphology through the morphology encoder. For each target hand, chain/joint correspondence to the Shadow reference is manually specified from its URDF according to the kinematic chains and joint ordering, allowing articulated hands with varying DoFs, chain counts, and base structures to be represented as long as their kinematic chains are serial. As illustrated in Fig.~\ref{fig:reference_mapping}, missing joints or chains are mapped to inactive reference slots, while extra joints or chains beyond the reference are fixed. Parallel or hybrid closed-chain mechanisms and hands without discrete articulated chains are outside the current scope.

A reference hand pose is represented as
$\boldsymbol{H}^c=(\boldsymbol{t},\boldsymbol{R},\boldsymbol{\theta}_c,\boldsymbol{\delta})\in\mathbb{R}^{9+2N_c}$,
where $N_c=24$, $\boldsymbol{\theta}_c\in\mathbb{R}^{N_c}$ denotes the reference joint angles, and $\boldsymbol{\delta}\in\{0,1\}^{N_c}$ indicates active joints. Missing reference slots are masked by setting $\delta_i=0$, with their joint and morphology features set to zero. We then formulate grasp generation as sampling
$P(\boldsymbol{H}^c|\boldsymbol{O},\mathcal{M})$.
After generation, the active reference joints are mapped back to the target hand through the same correspondence.

\noindent \textbf{3.3. Morphology Encoder} \label{sec:morph_encoder}

\begin{wrapfigure}{r}{0.4\linewidth}
    \centering
    \vskip -3em
    \includegraphics[width=\linewidth]{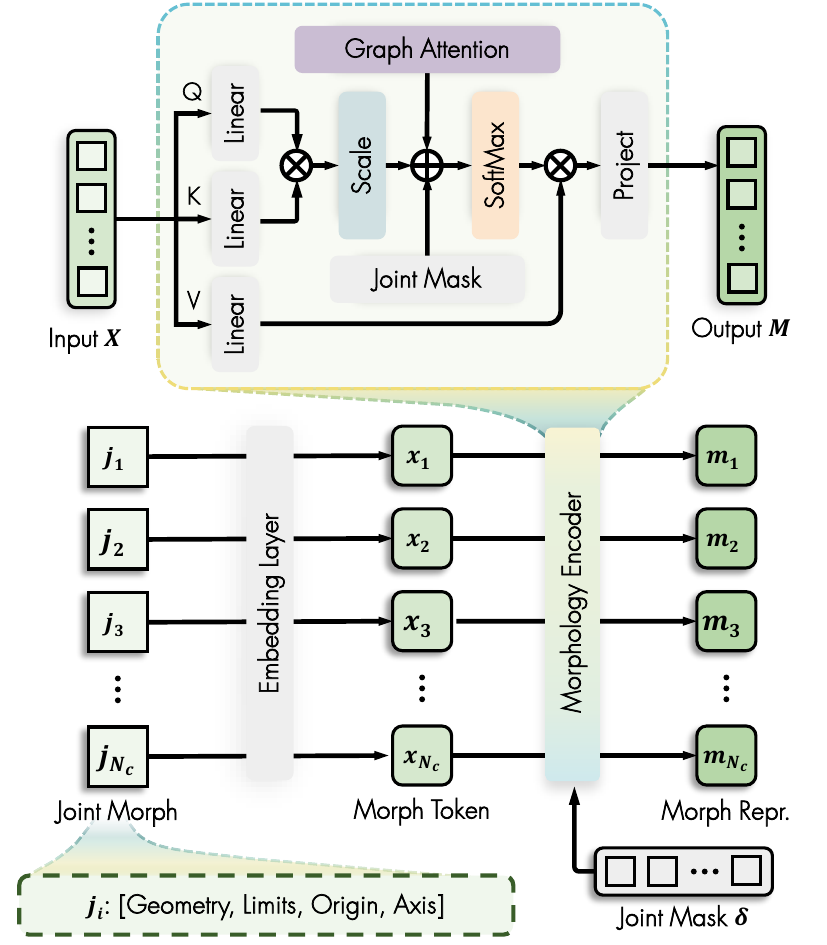}
    \caption{
    Morphology encoder.
    }
    \vskip -1.2em
    \label{fig:morph_enc}
\end{wrapfigure}

Inspired by ~\cite{patel2025getzero, ying2021graphormer}, we employ a morphology encoder to encode the hand's morphological information as graph-structured features. As shown in Fig.~\ref{fig:morph_enc}, we extract each joint's child link geometric properties, \textit{i.e.}, length, width, and height of the bounding box (3 dim), along with the joint's maximum and minimum limits (2 dim), origin (3 dim), and axis (3 dim), and concatenate them to obtain joint morphologies $\boldsymbol{J} = \{ \boldsymbol{j}_1, \dots, \boldsymbol{j}_{N_c} \} \in \mathbb{R}^{N_c \times 11}$. The morphs of each joint are embedded to obtain a set of morph tokens $\boldsymbol{X} = \{\boldsymbol{x}_1, \dots, \boldsymbol{x}_{N_c}\} \in \mathbb{R}^{N_c \times D}$, where $D$ denotes the feature dimension. The morphology encoder processes these tokens through self-attention layers with learned graph-based attention biases derived from the kinematic structure, including spatial adjacency and parent-child relationships, to obtain morph representations $\boldsymbol{M} = \{\boldsymbol{m}_1, \dots, \boldsymbol{m}_{N_c}\} \in \mathbb{R}^{N_c \times D}$. 
Specifically, the attention scores $A_{ij}$ between joints $i$ and $j$ are computed as:
\begin{equation}
A_{ij} = \frac{(\boldsymbol{x}_i \boldsymbol{w}_q)(\boldsymbol{x}_j \boldsymbol{w}_k)^T}{\sqrt{D}} + b_{\text{graph}}(i,j) + b_{\text{mask}}(\delta_i, \delta_j),
\end{equation}
where $\boldsymbol{w}_q, \boldsymbol{w}_k \in \mathbb{R}^{D \times D}$ are learnable projection matrices. The graph bias term $b_{\text{graph}}(i,j)$ is computed by passing the spatial adjacency matrix, parent matrix, and child matrix through separate embedding layers to encode the kinematic structure. 
The mask bias $b_{\text{mask}}(\delta_i, \delta_j)$ is obtained from the active joint mask defined in Sec.~\ref{sec:unified_hand_mapping}.
The final morph representation is computed as:
\begin{equation}
    \boldsymbol{M} = \text{softmax}(\boldsymbol{A})\boldsymbol{X}\boldsymbol{w}_v,
\end{equation}
where $\boldsymbol{w}_v \in \mathbb{R}^{D \times D}$ is a learnable projection matrix.

\noindent \textbf{3.4. Morphological Denoising Model} \label{sec:morph_unet}

We propose a morphological denoising model to continuously condition the iterative grasp generation on the morph representation $\boldsymbol{M}$, which enables the model to adapt grasp generation to diverse hand structures. 
As illustrated in Fig.~\ref{fig:pipeline} (Right), the conditioning inputs include the morph representation $\boldsymbol{M} \in \mathbb{R}^{N_c \times D}$ from Sec.~\ref{sec:morph_encoder} and the point cloud feature $\boldsymbol{P} \in \mathbb{R}^{N_p \times D}$ extracted from the object point cloud $\boldsymbol{O}$ via a Point Transformer~\cite{zhao2021pointtransformer} encoder, where $N_p$ represents the number of point groups. The initial hand feature $\boldsymbol{h}_T \in \mathbb{R}^{D}$ is obtained by embedding the noised hand pose $\boldsymbol{H}^c_T$ and the active joint mask $\boldsymbol{\delta}$ via MLPs, then concatenating them. The denoising process iteratively refines this feature from $t=T$ to $t=0$ through a series of Transformer blocks. At each timestep $t$, the current hand feature $\boldsymbol{h}_t$ is processed through a UNet block containing a residual convolutional layer conditioned on the timestep embedding: $\boldsymbol{h}_t^{\text{conv}} = \text{Conv}(\boldsymbol{h}_t) + \text{Emb}(t)$, where $\text{Emb}(t) \in \mathbb{R}^D$ that embeds timestep to the latent dimension. This is followed by cross-attention modules for morphology and point cloud conditioning. 
First, morphological conditioning is applied via cross-attention with $\boldsymbol{M}$:
\begin{equation}
\boldsymbol{h}_t^{m} = \text{softmax}\left(\frac{\boldsymbol{q}_h \boldsymbol{K}_M^T}{\sqrt{D}}\right)\boldsymbol{V}_M,
\end{equation}
where $\boldsymbol{q}_h = \boldsymbol{h}_t^{\text{conv}} \boldsymbol{w}_q \in \mathbb{R}^{1 \times D}$ is the query from the hand feature, and $\boldsymbol{K}_M = \boldsymbol{M} \boldsymbol{w}_k \in \mathbb{R}^{N_c \times D}$, $\boldsymbol{V}_M = \boldsymbol{M} \boldsymbol{w}_v \in \mathbb{R}^{N_c \times D}$ are the key and value from $\boldsymbol{M}$. Similarly, object-level conditioning is then applied via cross-attention with $\boldsymbol{P}$:
\begin{equation}
\boldsymbol{h}_t^{p} = \text{softmax}\left(\frac{\boldsymbol{q}'_h\boldsymbol{K}_P^T}{\sqrt{D}}\right)\boldsymbol{V}_P,
\end{equation}
where $\boldsymbol{q}'_h = \boldsymbol{h}_t^{m} \boldsymbol{w}_q' \in \mathbb{R}^{1 \times D}$ is the query from the morphology-conditioned hand feature, and $\boldsymbol{K}_P = \boldsymbol{P} \boldsymbol{w}_k' \in \mathbb{R}^{N_p \times D}$ and $\boldsymbol{V}_P = \boldsymbol{P} \boldsymbol{w}_v' \in \mathbb{R}^{N_p \times D}$ are derived from the grouped point cloud representation. The resulting feature $\boldsymbol{h}_t^{p}$ is then passed through a feedforward layer. By matching the single-token hand embedding with the multi-token graph-structured morph representation and grouped point cloud representation via cross-attention, our model effectively learns to generate grasps conditioned on the hand's kinematic and geometric structure. This process is repeated 8 times to iteratively denoise $\boldsymbol{h}_{t}$ and produce $\boldsymbol{h}_{t-1}$, ultimately obtaining $\boldsymbol{h}_0$, which is then decoded to generate the final grasp pose $\boldsymbol{H}^c_0$.

\noindent \textbf{3.5. Morphological Loss Function} \label{sec:morph_loss}

Due to the hierarchical nature of the hand kinematic tree, different joints have varying levels of influence on the overall pose. For instance, rotating a finger's base joint affects the entire finger chain (proximal, middle, and distal segments), whereas rotating the distal joint only impacts the fingertip. Motivated by this observation, we design a morphological loss function that adaptively scales joint errors based on their hierarchical positions in the kinematic tree using a non-linear geometric mean formulation, balancing proximal and distal gradients while enforcing kinematic constraints across diverse topologies during training. 

Given the kinematic structure of the hand, let $c_i$ denote the number of descendant joints for the $i$-th joint, where $c_i = 0$ for inactive joints. We compute an adaptive weight for each joint as:
\begin{equation} \label{eq:adaptive_weight}
    w_i = \sqrt{\frac{c_i + 1}{\mathcal{G}}},
\end{equation}
where the geometric mean $\mathcal{G}$ is defined as:
\begin{equation} \label{eq:geometric_mean}
    \mathcal{G} = \exp\left(\frac{1}{\sum_{j=1}^{N_c}\delta_j}\sum_{j=1}^{N_c} \delta_j \ln(c_j + 1)\right), 
\end{equation}
where $\boldsymbol{\delta} \in \mathbb{R}^{N_c}$ refers to the binary active joints mask. 
This formulation ensures that the weights maintain a geometric mean of 1, preventing overall scale drift while adaptively emphasizing proximal joints with larger kinematic influence. More details are provided in Appendix \ref{supp:geometric_mean}. 

The morphological loss is then formulated as:
\begin{equation}
    \mathcal{L}_m = \|\boldsymbol{t} - \hat{\boldsymbol{t}}\|_2^2 + \|\boldsymbol{R} - \hat{\boldsymbol{R}}\|_2^2 + \sum_{i=1}^{N_c} \delta_i w_i (\theta_i - \hat{\theta}_i)^2,
\end{equation}
where $(\boldsymbol{t}, \boldsymbol{R}, \boldsymbol{\theta})$ and $(\hat{\boldsymbol{t}}, \hat{\boldsymbol{R}}, \hat{\boldsymbol{\theta}})$ define $\boldsymbol{H}^c_0$ predicted by the morphological denoising model and the ground truth reference hand pose $\hat{\boldsymbol{H}^c_0}$, respectively. Following~\cite{wu2025cedex, zhong2025dexgraspanything}, we incorporate physics-aware losses to ensure generated grasps adhere to physical constraints. 
In detail, the total training loss is formulated as:
\begin{equation} \label{eq:loss_func}
    \mathcal{L} = \mathcal{L}_{\text{recon}} + \mathcal{L}_m + \alpha_{\text{spf}} \mathcal{L}_{\text{spf}} + \alpha_{\text{erf}} \mathcal{L}_{\text{erf}} + \alpha_{\text{srf}} \mathcal{L}_{\text{srf}},
\end{equation}
where $\mathcal{L}_{\text{recon}}$ denotes the standard noise reconstruction loss, and the physical constraint losses include the surface pulling force loss $\mathcal{L}_{\text{spf}}$, the external-penetration repulsion force loss $\mathcal{L}_{\text{erf}}$, and the self-penetration repulsion force loss $\mathcal{L}_{\text{srf}}$~\cite{zhong2025dexgraspanything}, which are detailed in Appendix \ref{supp:physical_constraints}. $\alpha_{\text{spf}}$, $\alpha_{\text{erf}}$, and $\alpha_{\text{srf}}$ are the corresponding balancing weights.

\section{Experiments}
\noindent \textbf{4.1. Implementation Details and Datasets}

We implement our model using PyTorch \cite{paszke2019pytorch} and conduct all experiments on a single 80GB NVIDIA A100 GPU. Following \cite{huang2023scenediffuser}, we transform the global hand pose $\{\boldsymbol{t}, \boldsymbol{R}, \boldsymbol{\theta}\}$ into $\{\boldsymbol{t}, \boldsymbol{\theta}\}$ by rotating the object point cloud via $\boldsymbol{R}^{-1}$, thereby aligning the hand pose with the object frame. We use Adam optimizer \cite{kingma2014adam} with a learning rate of 1e-4 and a batch size of 128 for 1 million iterations. We train on 14,011 grasps over 48 objects from the MultiDex training split \cite{li2023gendexgrasp}, covering Barrett, Allegro, and Shadow hands. For evaluation, in addition to MultiDex test set, we perform cross-dataset evaluations on Multi-GraspLLM \cite{li2024multigraspllm} and Objaverse \cite{deitke2023objaverse} datasets, following GraspXL \cite{zhang2024graspxl} and CEDex \cite{wu2025cedex}, to assess zero-shot cross-dataset generalization. We sample 64 grasps for each evaluation object.

\noindent \textbf{4.2. Baselines and Evaluation Metrics}

We take established cross-embodiment dexterous grasp synthesis methods as our baselines, including optimization-based approaches such as DFC \cite{liu2021dfc}, GenDexGrasp \cite{li2023gendexgrasp}, and CEDex \cite{wu2025cedex}, as well as learning-based methods including GeoMatch series \cite{attarian2023geomatch, wei2024geomatch++}, and DRO-Grasp \cite{wei2024drograsp}. To evaluate grasp synthesis performance, we employ success rate, diversity, and efficiency as our main metrics. We also evaluate penetration depth and Q1 metric. The details of metrics are provided in Appendix \ref{supp:metrics}.

\begin{table*}[t!]  
    \centering  
    \caption{Quantitative results of our DiffMorphGrasp (w/. and w/o. the morphological loss) compared with cross-embodiment dexterous grasp synthesis baselines on the MultiDex \cite{li2023gendexgrasp} dataset across three robotic hands: Barrett, Allegro, and Shadow hands. }  
    \label{tab:multidex_comparison}  
    \fontsize{7.5pt}{9pt}\selectfont
    \begin{tabular}{l | c c c c | c c c c | c}
        \toprule  
        \multirow{2}{*}{Method} & \multicolumn{4}{c}{Success Rate (\%) $\uparrow$} & \multicolumn{4}{c}{Diversity (rad.) $\uparrow$} & \multirow{2}{*}{Eff. (s) $\downarrow$}\\
        \cmidrule(lr){2-5} \cmidrule(lr){6-9}
        & Barrett & Allegro & ShadowHand & \textbf{Avg.} & Barrett & Allegro & ShadowHand & \textbf{Avg.} \\
        \midrule
        DFC 
        & 86.3 & 76.2 & 58.8 & 73.8 & 0.532 & 0.454 & 0.435 & 0.474 & $>$1800 \\
        GenDexGrasp 
        & 67.0 & 51.0 & 54.2 & 57.4 & 0.488 & 0.389 & 0.318 & 0.398 & 19.7 \\
        CEDex 
        & \cellcolor{LimeGreen}93.1 & 88.1 & 85.0 & 88.7 & 0.624 & \cellcolor{LimeGreen}0.473 & 0.438 & 0.512 & 7.8 \\
        GeoMatch 
        & 60.0 & - & 67.5 & 63.8 &  0.259 & - &  0.235 & 0.247 & - \\
        GeoMatch++ 
        & 77.5 & - & 70.0 & 73.8 &  0.378 & - &  0.184 & 0.281 & - \\
        DRO-Grasp 
        & 87.3 & \cellcolor{LimeGreen}\textbf{92.3} & 83.0 & 87.5 & 0.513 & 0.397 & 0.441 & 0.450 & 
        0.65 \\
        \hline
        \textbf{Ours} w/o. $\mathcal{L}_m$ & 92.5 & 89.6 & \cellcolor{LimeGreen!50}95.0 & \cellcolor{LimeGreen!50}92.4 & \cellcolor{LimeGreen!50}0.696 & 0.434 & \cellcolor{LimeGreen}\textbf{0.451} & \cellcolor{LimeGreen!50}0.527 & \cellcolor{LimeGreen}\textbf{0.47} \\
        \textbf{Ours} w/. $\mathcal{L}_m$ & \cellcolor{LimeGreen!50}\textbf{93.0} & \cellcolor{LimeGreen!50}90.3 & \cellcolor{LimeGreen}\textbf{98.8} & \cellcolor{LimeGreen}\textbf{94.0} & \cellcolor{LimeGreen}\textbf{0.698} & \cellcolor{LimeGreen!50}\textbf{0.462} & \cellcolor{LimeGreen!50}0.445 & \cellcolor{LimeGreen}\textbf{0.535} & \cellcolor{LimeGreen}\textbf{0.47} \\
        \bottomrule  
    \end{tabular}  
    \vskip -1em
\end{table*}


\begin{figure*}[t!]
    \centering
    \includegraphics[width=\linewidth]{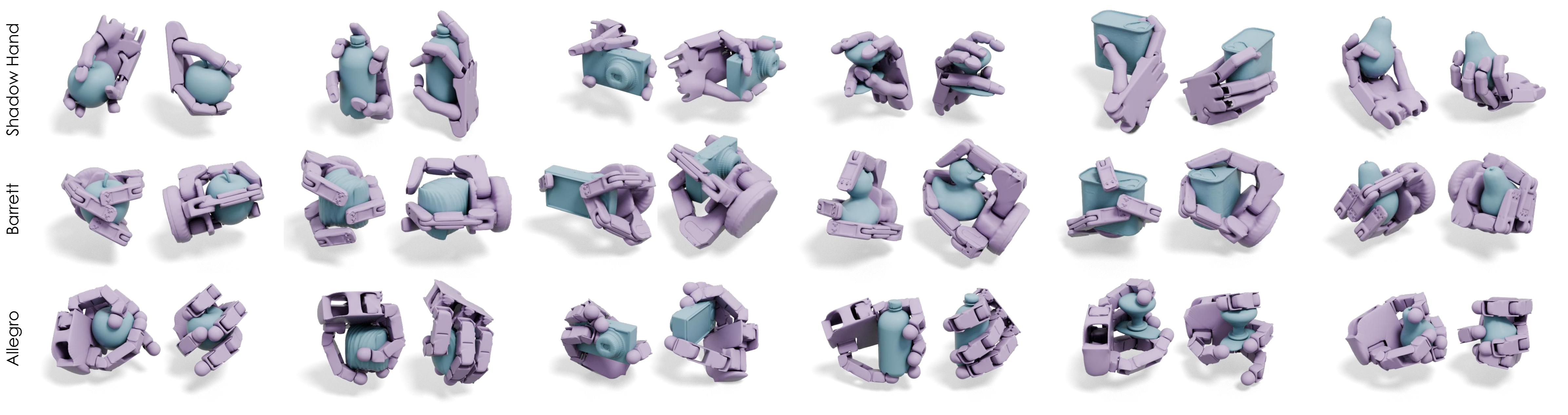}
    \caption{ 
        Visualizations of cross-embodiment grasps synthesized by DiffMorphGrasp.
    }
    \label{fig:vis_multidex}
    \vskip -1em
\end{figure*}

\noindent \textbf{4.3. Comparison with SoTA}

We present the quantitative results on the MultiDex dataset \cite{li2023gendexgrasp} in Tab. \ref{tab:multidex_comparison}. Qualitative results are visualized in Fig. \ref{fig:vis_multidex} and Appendix Figs. \ref{fig:vis_more_multidex} and \ref{fig:qualitative_comparison} for visual comparisons against baselines. Zero-shot cross-dataset evaluation (out of domain) results are provided in Appendix \ref{supp:cross_dataset}. 

As shown in Tab. \ref{tab:multidex_comparison}, CEDex and DRO-Grasp achieve the highest success rate among the baselines. DiffMorphGrasp outperforms all evaluated baselines across all metrics, achieving an overall success rate of 92.4\%, representing improvements ranging from 3.7\% to 35.0\% over the baselines. It also demonstrates enhanced efficiency with an average execution time of 0.47 seconds per grasp. Notably, our model achieves the highest diversity of 0.527, corresponding to a large increase compared to the baselines. This enhanced diversity stems from the inherent characteristics of diffusion models in producing variable results. Importantly, our morphology encoding provides a richer understanding of hand structures, enabling it to generate more stable and diverse grasps. This is particularly evident with the Shadow hand where DiffMorphGrasp substantially improves success rates by 10.0\%. Adding the morphological loss function further results in an improvement of 1.6\% in overall success rate and 1.5\% in diversity.

\noindent \textbf{4.4. Zero-Shot Generalization Ability}

\begin{figure*}[t!]
    \centering
    \includegraphics[width=\linewidth]{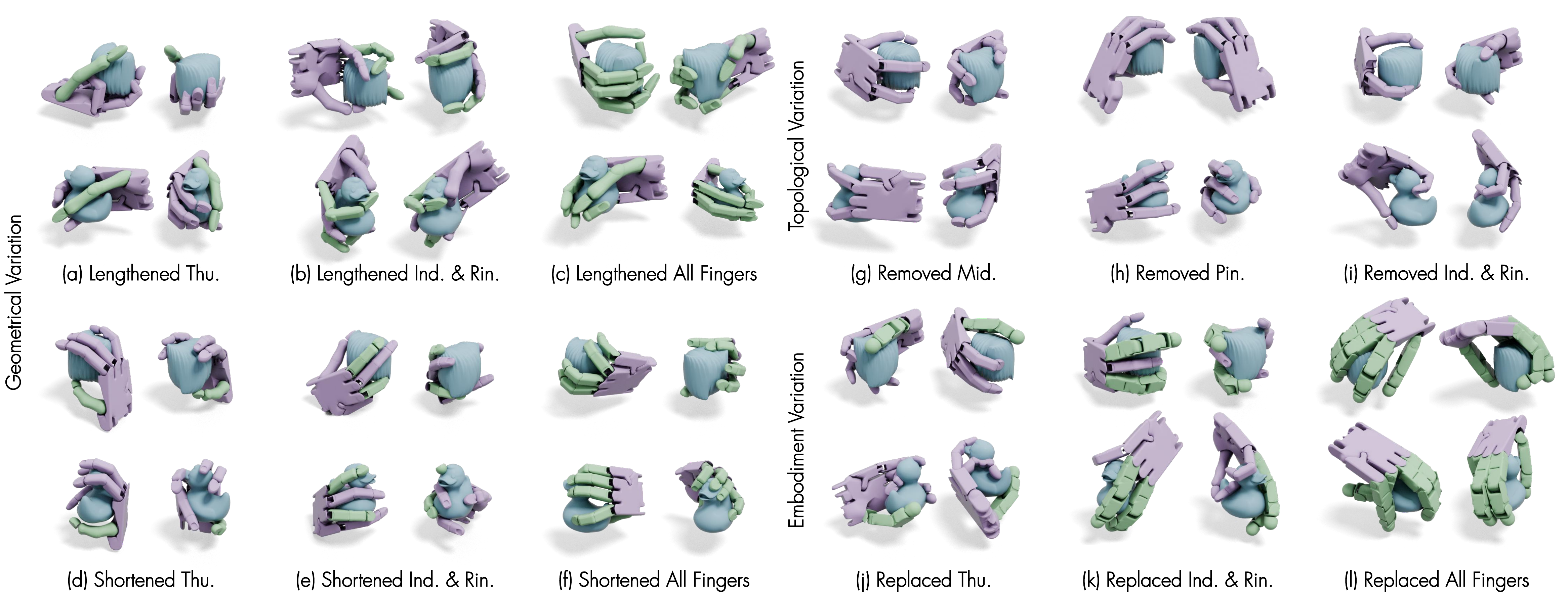}
  \vskip -0.4em
    \caption{ 
        Zero-shot generalization to structural variations of the Shadow hand. (a)-(f) geometrical variations: lengthen and shorten fingers. (g)-(i) topological variations: remove fingers. (j)-(l) hybrid variations: replace fingers with Allegro. Altered fingers are highlighted in green. 
    }
    \label{fig:vis_generalize}
\end{figure*}

\begin{figure*}[t!]
    \centering
    \vskip -1.2em
    \includegraphics[width=\linewidth]{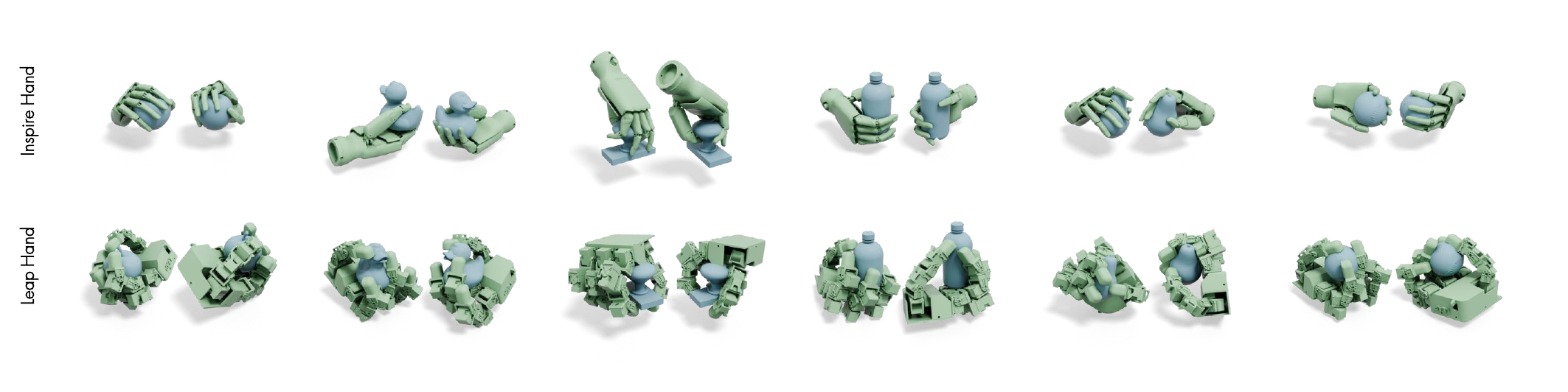}
    \vskip -1.6em
    \caption{ 
        Visualizations of zero-shot grasp generalization to unseen Inspire and Leap hands. Green highlights that the morphologies are completely unseen during training. 
    }
    \label{fig:unseen_more}
\end{figure*}

An important advantage of our diffusion-based framework is zero-shot generalization to unseen hand structures. To validate this, we evaluate two settings: (1) modified Shadow hands with topological, geometrical, and hybrid embodiment variations, and (2) completely unseen Leap and Inspire hands with different kinematic structures. Qualitative results on modified Shadow hands are shown in Fig. \ref{fig:vis_generalize}, with quantitative and more qualitative results in Appendix \ref{supp:more_generalize}. Quantitative results of completely unseen hands are provided in Tab. \ref{tab:unseen}, with qualitative results in Fig. \ref{fig:unseen_more}. While DRO-Grasp fails on Inspire hand with only a 6.9\% success rate, our model achieves high success rates (Inspire hand 86.3\% and Leap hand 83.7\%), as well as robust diversity, penetration depth, and Q1 metrics. These results demonstrate our model's capability to generalize to novel hand structures without retraining. 

\begin{table}[t!]
    \centering
    \vskip -0.4em
    \caption{
    Quantitative results of zero-shot generalization to completely unseen hand morphologies, \textit{i.e.}, Inspire and Leap hands on the MultiDex \cite{li2023gendexgrasp} dataset. We report success rate, diversity, penetration depth, and Q1 metric. 
    }
    \label{tab:unseen}
    \fontsize{7.5pt}{9pt}\selectfont
    \begin{tabular}{l c c c c}
        \toprule
        Methods & Suc. Rat. (\%) $\uparrow$ & Div. (rad) $\uparrow$ & Pen. Dep. (mm) $\downarrow$ & Q1 Metric $\uparrow$ \\
        \midrule
        Inspire - DRO-Grasp & 6.9 & \textbf{0.636} & 20.88 & 0.0041 \\
        Inspire - \textbf{Ours} & \textbf{86.3} & 0.466 & \textbf{10.02} & \textbf{0.0281} \\
        \hdashline
        Leap - DRO-Grasp & 43.8 & 0.353 & 22.03 & 0.0203 \\
        Leap - \textbf{Ours} & \textbf{83.7} & \textbf{0.409} & \textbf{12.28} & \textbf{0.0294} \\
        \bottomrule
    \end{tabular}
    \vskip -0.4em
\end{table}

\newpage

\begin{figure*}[t!]
    \centering
    \includegraphics[width=\linewidth]{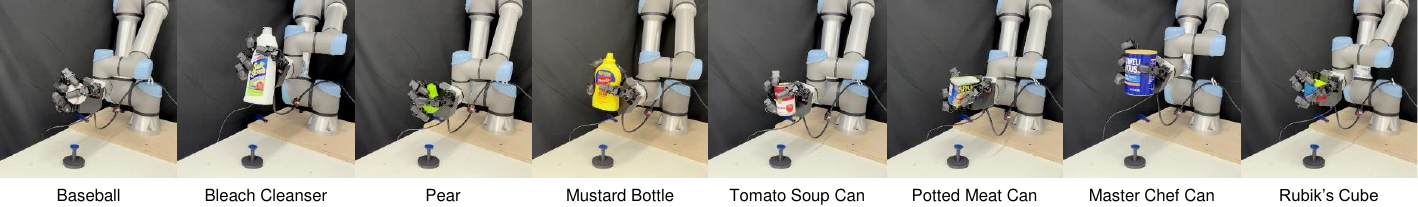}
    \vskip -0.4em
    \caption{ 
        Real-world validation using the Leap Hand \cite{shaw2023leaphand} demonstrates stable grasps on eight YCB objects \cite{calli2015ycb}. Video results are provided in our supplementary demo.
    }
    \label{fig:real_world}
    \vskip -0.4em
\end{figure*}

\noindent \textbf{4.5. Real-World Validation}

\begin{table}[t!]
    \centering
    \caption{
    Quantitative real-world evaluation on the Leap Hand \cite{shaw2023leaphand}. We report the success rate over 10 attempts for eight objects from the YCB dataset \cite{calli2015ycb}.
    }
    \label{tab:real_world}
    \fontsize{7.5pt}{9pt}\selectfont
    \begin{tabular}{cccccccc}
        \toprule
        Baseball & Cleanser & Pear & Mustard Bottle & Soup Can & Meat Can & Chef Can & Rubik's Cube  \\
        \midrule
        9/10 & 10/10 & 10/10 & 8/10 & 9/10 & 8/10 & 10/10 & 9/10  \\
        \bottomrule
    \end{tabular}
    \vskip -1.2em
\end{table}

We validate DiffMorphGrasp in real-world scenarios using a UR5e arm equipped with a Leap Hand \cite{shaw2023leaphand}. We conduct 10 attempts on each of eight YCB objects \cite{calli2015ycb}. As reported in Tab. \ref{tab:real_world}, our approach achieves an overall success rate of 91\% (73/80), demonstrating robust performance across diverse geometries. Qualitative examples in Fig. \ref{fig:real_world} further confirm the practical effectiveness of our method. More details and additional real-world experiments are provided in Appendices \ref{supp:real_world} and \ref{supp:closed_loop}.

\noindent \textbf{4.6. Failure Cases} \label{sec:failure_cases}

\begin{figure*}[t!]
    \centering
    \includegraphics[width=0.6\linewidth]{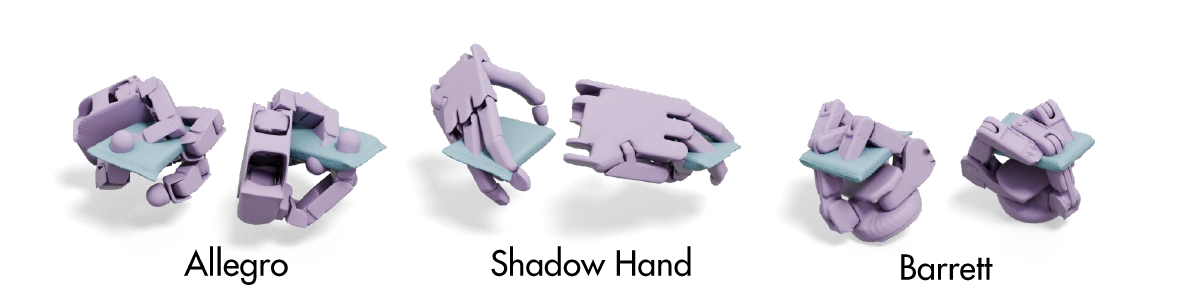}
    \vskip -0.4em
    \caption{ 
        Failure cases of DiffMorphGrasp. We visualize grasp attempts on a thin sponge across three different robotic hands, which illustrate the limitation of object-hand penetration. 
    }
    \label{fig:failure_cases}
    \vskip -0.4em
\end{figure*}

The main failure mode for our model is object-hand penetration, particularly for thin objects, as shown in Fig. \ref{fig:failure_cases}. We attribute this mainly to two factors. First, the training grasps prioritize grasp stability rather than enforcing collision-free configurations. Second, the surface pulling loss encourages the hand surface to approach the object for stable contact, which can lead to over-tight grasps and penetration for thin geometries.

\section{Conclusion}
In this paper, we introduced \textbf{DiffMorphGrasp}, the first diffusion-based framework that incorporates hand morphology information into the generative process for cross-embodiment dexterous grasping, which enables zero-shot generalization to unseen hand structures. Our approach continuously conditions the generative process with graph-based hand morphology features, with a morphological loss function that adaptively enforces hierarchical joint relationships. Extensive experiments demonstrate that DiffMorphGrasp achieves state-of-the-art performance on multiple benchmarks and generalizes effectively to novel hand structures in a zero-shot manner.

\section{Limitations and Future Work}
\label{sec:limitation}

A limitation of the current framework is that grasp generation does not use physical contact feedback during execution, leaving residual object-hand penetration unresolved in some cases (Fig. \ref{fig:failure_cases}). Incorporating tactile sensing could enable online pose refinement using real contact feedback and mitigate penetration \cite{wu2019mat}. Another limitation is inference frequency. Although our 10-step DPM-Solver achieves 0.47\,s inference, diffusion distillation could further reduce the number of denoising steps and improve inference frequency for more reactive grasp generation.

\acknowledgments{This work was supported in part by the EPSRC project ``TacDiff: Designing Tactile-based Robots via Differentiable Simulations'' (UKRI3438).}


\bibliography{main}

\appendix
\onecolumn
\section*{Appendix}

\section{Loss Function} \label{supp:physical_constraints}
As formulated in Eq.~\ref{eq:loss_func}, the total training objective comprises a reconstruction loss, a morphological loss, and physical constraint losses. Consistent with standard diffusion models, the reconstruction loss $\mathcal{L}_{\text{recon}}$ is defined as the mean squared error between the sampled Gaussian noise $\boldsymbol{\epsilon}$ and the noise predicted by the network $\boldsymbol{\epsilon}_\phi$:
\begin{equation}
    \mathcal{L}_{\text{recon}} = \|\boldsymbol{\epsilon} - \boldsymbol{\epsilon}_\phi(\boldsymbol{H}^c_t, \boldsymbol{O}, \mathcal{M}, t)\|_2^2,
\end{equation}
where $\boldsymbol{\epsilon} \sim \mathcal{N}(\mathbf{0}, \mathbf{I})$ denotes the sampled Gaussian noise. The term $\boldsymbol{\epsilon}_\phi$ represents the denoising network, which predicts the noise component given the noisy reference hand pose $\boldsymbol{H}^c_t$ at timestep $t$.

To ensure physical plausibility and grasp stability, we follow~\cite{zhong2025dexgraspanything, xu2023unidexgrasp} and incorporate three auxiliary physical loss terms. These constraints are evaluated on the sampled surface points $\mathcal{P}$ of the hand mesh, reconstructed from the predicted reference hand pose $\boldsymbol{H}^c_0$. Specifically, we employ: (1) Surface Pulling Force (SPF) loss~\cite{xu2023unidexgrasp} to encourage contact between the hand and the object surface; (2) External-Penetration Repulsion Force (ERF) loss~\cite{li2023gendexgrasp} to penalize collisions between the hand and the object; and (3) Self-Penetration Repulsion Force (SRF) loss~\cite{xu2023unidexgrasp} to prevent physically impossible self-intersections of the hand links. These losses are formulated as:

\begin{equation}
\mathcal{L}_{\text{spf}} = \frac{1}{|\mathcal{S}| + \epsilon} \sum_{\boldsymbol{p} \in \mathcal{S}} \sqrt{d(\boldsymbol{p}, \boldsymbol{O})},
\end{equation}

\begin{equation}
\mathcal{L}_{\text{erf}} = \frac{1}{|\mathcal{P}|} \sum_{\boldsymbol{p} \in \mathcal{P}} \max\left(0, -\text{SDF}_{\boldsymbol{O}}(\boldsymbol{p})\right),
\end{equation}
and
\begin{equation}
\mathcal{L}_{\text{srf}} = \frac{1}{N_{link}} \sum_{l=1}^{N_{link}} \sum_{\boldsymbol{p}_{i,j} \in \mathcal{P}_l, i \neq j} \max(0, d_{th} - \|\boldsymbol{p}_i - \boldsymbol{p}_j\|_2),
\end{equation}
where $d(\boldsymbol{p}, \boldsymbol{O})$ denotes the Euclidean distance from a hand surface point $\boldsymbol{p}$ to the object point cloud $\boldsymbol{O}$, and $\mathcal{S} = \{ \boldsymbol{p} \in \mathcal{P} \mid d(\boldsymbol{p}, \boldsymbol{O}) < \tau \}$ represents the set of hand points within a proximity threshold $\tau$. $\text{SDF}_{\boldsymbol{O}}(\boldsymbol{p})$ is the signed distance function of the object evaluated at point $\boldsymbol{p}$. For self-penetration, we compute the pairwise distances between points belonging to different links, where $\mathcal{P}_l$ denotes the set of points on the $l$-th link, and $d_{th}$ is the collision threshold.

\section{Evaluation Metrics} \label{supp:metrics}
Following established protocols \cite{li2023gendexgrasp, wei2024drograsp}, we employ success rate, diversity, and efficiency as our main metrics, defined as:

\begin{itemize}
    \item \textbf{Success Rate}: We evaluate grasping success by applying external forces to the object and measuring its displacement. Using Isaac Gym simulator \cite{liang2018isaacgym}, a simple grasp controller executes the predicted grasps \cite{wei2024drograsp}. Following metric definition in \cite{li2023gendexgrasp}, we sequentially apply forces along six orthogonal directions for 1 second each. A grasp is considered successful if the object's displacement remains below 2 cm once all forces are applied.
    \item \textbf{Diversity}: Grasp diversity is quantified by computing the standard deviation of joint configurations across all successful grasps, including the 6-DoF wrist pose and finger joint angles. Higher standard deviation indicates greater diversity in the generated grasp configurations.
    \item \textbf{Efficiency}: We assess efficiency by measuring the time it takes to generate each grasp pose. Lower execution times indicate a more efficient grasp synthesis process.
\end{itemize}

To provide a comprehensive performance evaluation, we also employ \textbf{penetration depth} and the \textbf{Q1 metric}, following \cite{wang2023dexgraspnet, liu2020ddg}, defined as the maximum hand-object penetration depth in millimeters and the force-closure grasp quality, respectively. Results are provided in Appendix \ref{supp:depth_penetration_q1}. 

\section{Depth Penetration and Q1 Metric}
\label{supp:depth_penetration_q1}

\begin{table*}[ht!]  
    \centering  
    \caption{
    Quantitative results of penetration depth and Q1 metric compared to cross-embodiment dexterous grasp synthesis baselines on the MultiDex \cite{li2023gendexgrasp} dataset. }  
    \label{tab:penetration_depth_q1}
    \fontsize{7.5pt}{9pt}\selectfont
    \begin{tabular}{l | c c c c | c c c c}
        \toprule  
         \multirow{2}{*}{Methods} & \multicolumn{4}{c}{Pen. Dep. (mm) $\downarrow$} & \multicolumn{4}{c}{Q1 Metric $\uparrow$} \\
        \cmidrule(lr){2-5} \cmidrule(lr){6-9}
        & Barrett & Allegro & ShadowHand & \textbf{Avg.} & Barrett & Allegro & ShadowHand & \textbf{Avg.} \\
        \midrule
        GenDexGrasp & 15.8 & 18.1 & 21.4 & 18.4 & \cellcolor{LimeGreen!50}0.0194 & \cellcolor{LimeGreen!50}0.0255 & \cellcolor{LimeGreen!50}0.0232 & \cellcolor{LimeGreen!50}0.0227 \\
        DRO-Grasp & \cellcolor{LimeGreen!50}14.0 & \cellcolor{LimeGreen!50}12.1 & \cellcolor{LimeGreen!50}13.9 & \cellcolor{LimeGreen!50}13.3 & 0.0059 & 0.0116 & 0.0149 & 0.0108 \\
        \textbf{Ours} & \cellcolor{LimeGreen}\textbf{9.3} & \cellcolor{LimeGreen}\textbf{10.9} & \cellcolor{LimeGreen}\textbf{8.5} & \cellcolor{LimeGreen}\textbf{9.6} & \cellcolor{LimeGreen}\textbf{0.0255} & \cellcolor{LimeGreen}\textbf{0.0300} & \cellcolor{LimeGreen}\textbf{0.0284} & \cellcolor{LimeGreen}\textbf{0.0280} \\
        \bottomrule  
    \end{tabular}  
\end{table*}

We provide quantitative results of depth penetration and Q1 metric in Tab. \ref{tab:penetration_depth_q1}. Compared to SoTA baselines, our DiffMorphGrasp achieves the best performance, improving penetration depth by 27.8\%-47.8\% and Q1 metric by 23.3\%-159.3\%.

\section{More Visualizations of Generated Grasps} \label{supp:more_viz}

\begin{figure*}[ht!]
    \centering
    \includegraphics[width=\linewidth]{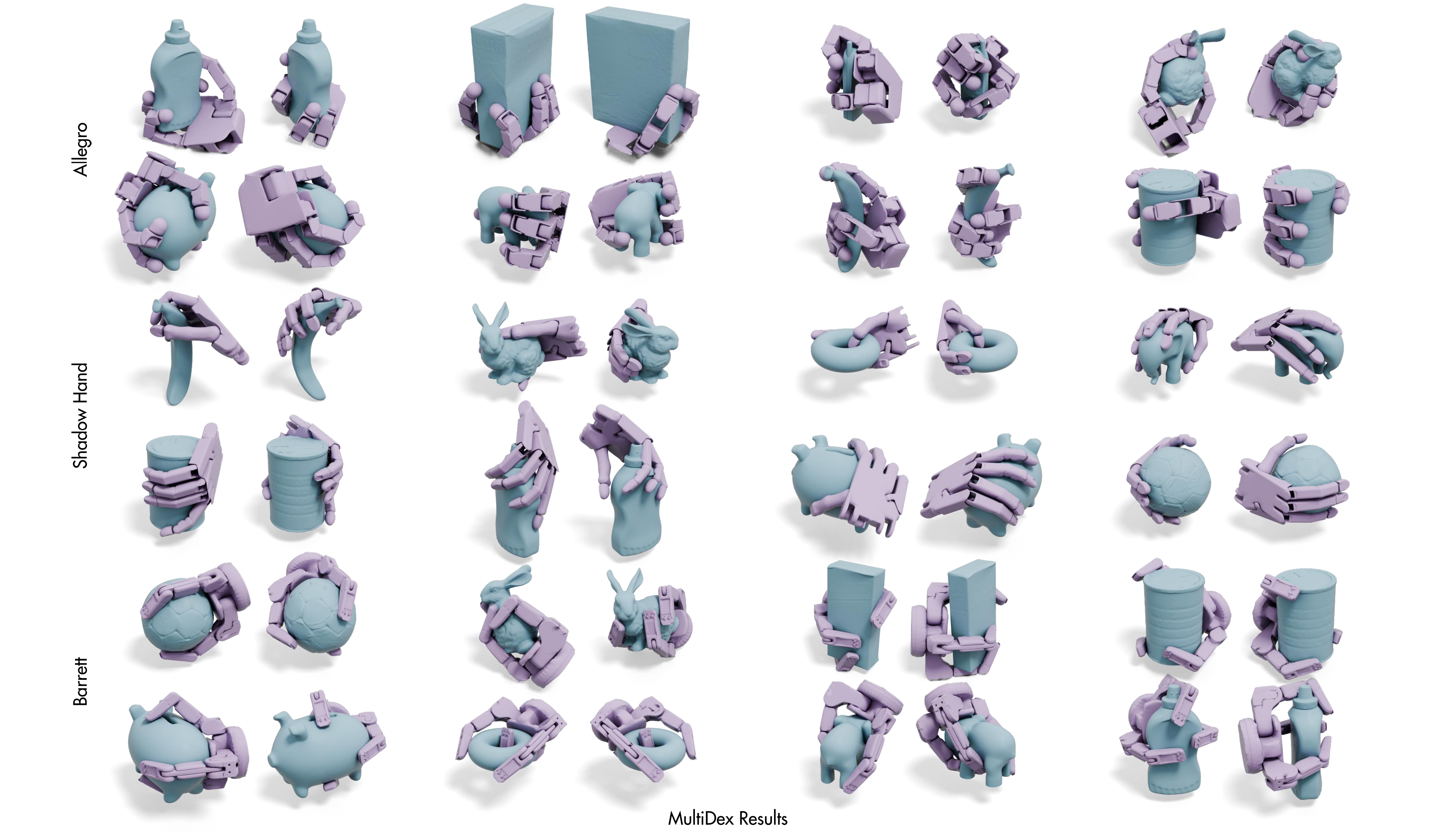}
    \vskip -0.6em
    \caption{ 
        Visualizations of cross-embodiment grasps synthesized by DiffMorphGrasp on the MultiDex \cite{li2023gendexgrasp} dataset. Two viewing angles are presented for each grasp.
    }
    \label{fig:vis_more_multidex}
\end{figure*}

We present extended qualitative results on the MultiDex dataset \cite{li2023gendexgrasp} in Fig. \ref{fig:vis_more_multidex}, demonstrating the robustness of DiffMorphGrasp in synthesizing physically stable and kinematically diverse grasps.

\section{Qualitative Comparison with Baselines} 
\label{supp:qualitative_comparison}

\begin{figure*}[ht!]
    \centering
    \includegraphics[width=\linewidth]{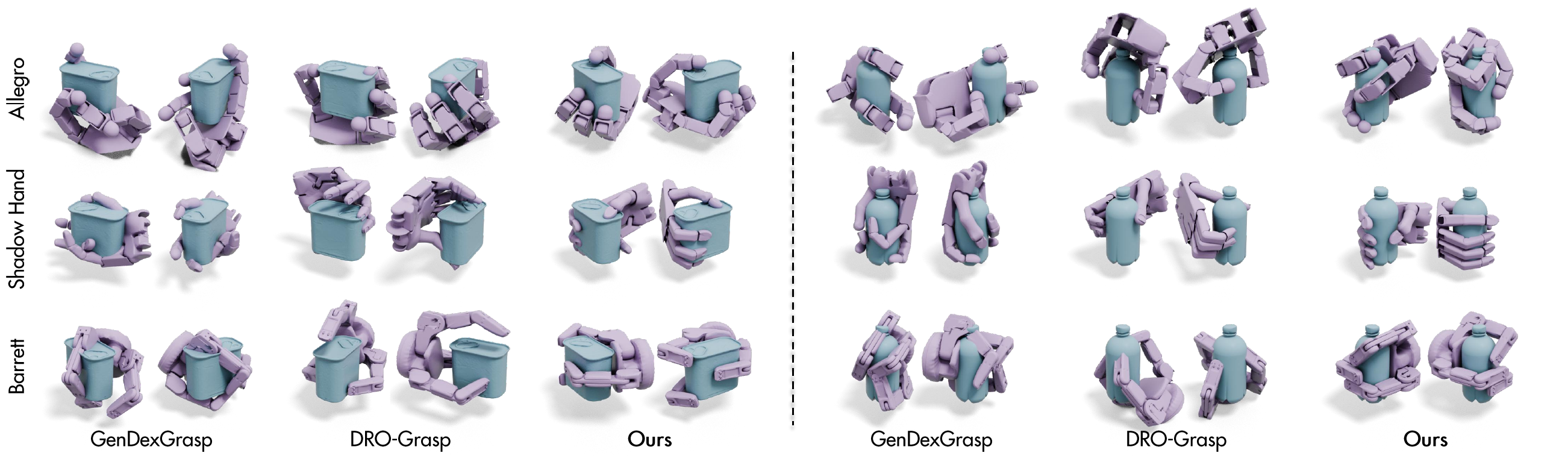}
    \vskip -0.6em
    \caption{ 
        Qualitative comparison with baselines 1) GenDexGrasp \cite{li2023gendexgrasp} and 2) DRO-Grasp \cite{wei2024drograsp} where our results demonstrate superior surface conformity and stable form-closure. Two viewing angles are presented for each grasp.
    }
    \label{fig:qualitative_comparison}
\end{figure*}

We provide qualitative comparison with baselines with GenDexGrasp \cite{li2023gendexgrasp} and DRO-Grasp \cite{wei2024drograsp} in Fig. \ref{fig:qualitative_comparison}. GenDexGrasp frequently exhibits physically infeasible object penetration, whereas DRO-Grasp tends to generate loose configurations that lack sufficient contact and heavily rely on downstream controllers for execution. In contrast, our method synthesizes stable, tight-fitting grasps with high surface conformity across different embodiments.

\section{Out-of-Domain Zero-Shot Cross-Dataset Evaluation} \label{supp:cross_dataset}

\begin{figure*}[ht!]
    \centering
    \includegraphics[width=\linewidth]{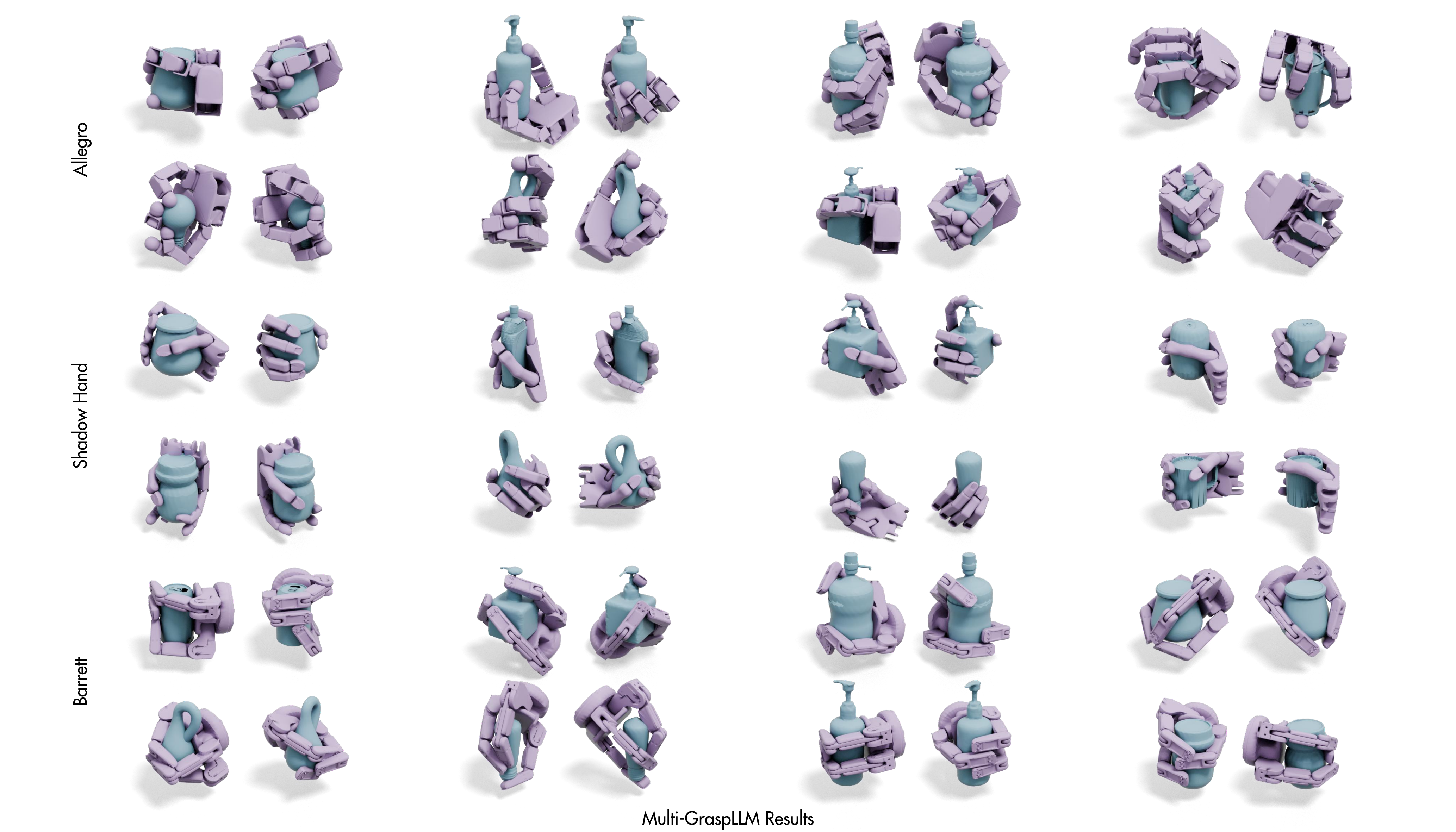}
    \vskip -0.6em
    \caption{ 
        Visualizations of cross-embodiment grasps synthesized by DiffMorphGrasp on the Multi-GraspLLM \cite{li2024multigraspllm} dataset. Two viewing angles are presented for each grasp.
    }
    \label{fig:vis_more_multigraspset}
\end{figure*}

\begin{figure*}[ht!]
    \centering
    \includegraphics[width=\linewidth]{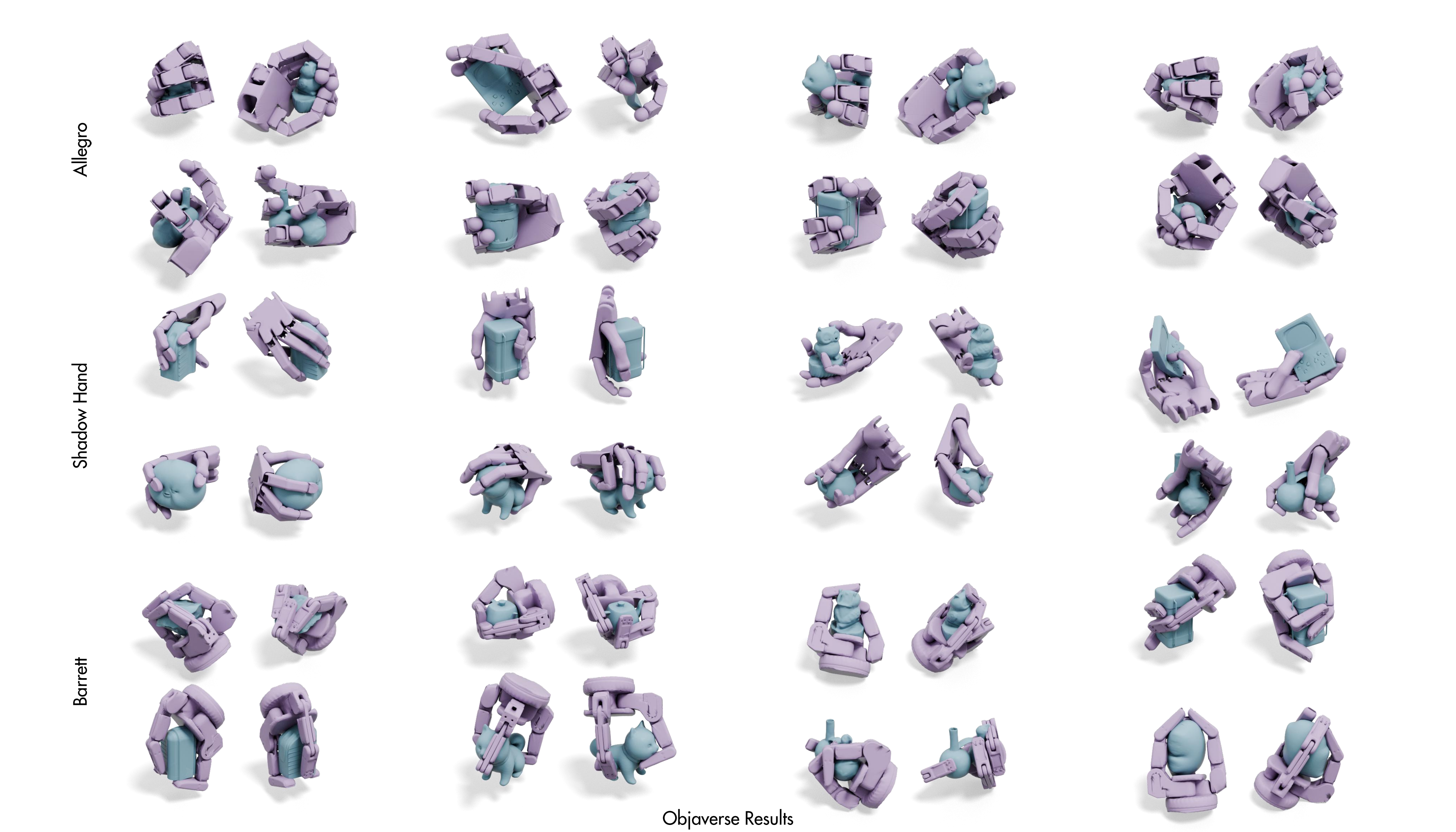}
    \vskip -0.6em
    \caption{ 
        Visualizations of cross-embodiment grasps synthesized by DiffMorphGrasp on the Objaverse \cite{deitke2023objaverse} dataset. Two viewing angles are presented for each grasp.
    }
    \label{fig:vis_more_objaverse}
\end{figure*}

Tab. \ref{tab:cross_dataset} presents the results of zero-shot cross-dataset evaluation on the Multi-GraspLLM \cite{li2024multigraspllm} and Objaverse \cite{deitke2023objaverse} datasets. Complementing this, qualitative results are visualized in Figs. \ref{fig:vis_more_multigraspset} and \ref{fig:vis_more_objaverse}. 

\begin{table*}[t!]
    \centering
    \caption{Cross-dataset zero-shot generalization results. We evaluate models trained on MultiDex \cite{li2023gendexgrasp} directly on unseen datasets: Multi-GraspLLM \cite{li2024multigraspllm} and Objaverse \cite{deitke2023objaverse}. }
    \label{tab:cross_dataset}
    \fontsize{7.5pt}{9pt}\selectfont
    \begin{tabular}{l l | c c c c | c c c c}
        \toprule
        \multirow{2}{*}{Test Dataset} & \multirow{2}{*}{Method} & \multicolumn{4}{c}{Success Rate (\%) $\uparrow$} & \multicolumn{4}{c}{Diversity (rad.) $\uparrow$} \\
        \cmidrule(lr){3-6} \cmidrule(lr){7-10}
        & & Barrett & Allegro & Shadow & \textbf{Avg.} & Barrett & Allegro & Shadow & \textbf{Avg.} \\
        \midrule
        \multirow{3}{*}{\shortstack[l]{Multi-GraspLLM}} 
        & GenDexGrasp 
        & 29.3 & 38.9 & 46.7 & 38.3 & \cellcolor{LimeGreen!50}0.620 & \cellcolor{LimeGreen!50}0.398 & 0.409 & \cellcolor{LimeGreen!50}0.476 \\
        & DRO-Grasp 
        & \cellcolor{LimeGreen!50}81.1 & \cellcolor{LimeGreen!50}82.7 & \cellcolor{LimeGreen!50}71.4 & \cellcolor{LimeGreen!50}78.4 & 0.512 & 0.395 & \cellcolor{LimeGreen!50}0.458 & 0.455 \\
        & \textbf{Ours} 
        & \cellcolor{LimeGreen}\textbf{84.7} & \cellcolor{LimeGreen}\textbf{84.9} & \cellcolor{LimeGreen}\textbf{92.5} & \cellcolor{LimeGreen}\textbf{87.4} & \cellcolor{LimeGreen}\textbf{0.708} & \cellcolor{LimeGreen}\textbf{0.470} & \cellcolor{LimeGreen}\textbf{0.463} & \cellcolor{LimeGreen}\textbf{0.547} \\
        \midrule
        \multirow{3}{*}{Objaverse} 
        & GenDexGrasp 
        & 57.9 & 42.0 & 63.9 & 54.6 & \cellcolor{LimeGreen!50}0.597 & \cellcolor{LimeGreen!50}0.436 & 0.398 & \cellcolor{LimeGreen!50}0.477 \\
        & DRO-Grasp 
        & \cellcolor{LimeGreen!50}82.2 & \cellcolor{LimeGreen!50}82.9 & \cellcolor{LimeGreen!50}70.3 & \cellcolor{LimeGreen!50}78.5 & 0.517 & 0.396 & \cellcolor{LimeGreen!50}0.445 & 0.453 \\
        & \textbf{Ours} 
        & \cellcolor{LimeGreen}\textbf{89.9} & \cellcolor{LimeGreen}\textbf{91.1} & \cellcolor{LimeGreen}\textbf{92.9} & \cellcolor{LimeGreen}\textbf{91.3} & \cellcolor{LimeGreen}\textbf{0.728} & \cellcolor{LimeGreen}\textbf{0.451} & \cellcolor{LimeGreen}\textbf{0.448} & \cellcolor{LimeGreen}\textbf{0.541} \\
        
        \bottomrule
    \end{tabular}
\end{table*}

For evaluation, we select a test set of 20 representative objects for each dataset. On Multi-GraspLLM, existing baselines struggle to balance performance metrics. GenDexGrasp shows high diversity but suffers from a low success rate of 38.3\%, whereas DRO-Grasp achieves a 78.4\% success rate but with limited diversity of 0.455 rad. In contrast, DiffMorphGrasp simultaneously achieves state-of-the-art performance in both stability with 87.4\% success rate and diversity with 0.547 rad, surpassing baselines significantly. On the Objaverse dataset, our model achieves an average success rate of 91.3\%, outperforming the second-best method DRO-Grasp by 12.8\%. This advantage is particularly pronounced for complex kinematic structures. For instance, on the high-DoF Shadow hand within Multi-GraspLLM, baseline performance degrades notably with GenDexGrasp dropping to 46.7\% and DRO-Grasp to 71.4\%, whereas DiffMorphGrasp maintains a robust 92.5\% success rate. The results demonstrate that our proposed DiffMorphGrasp effectively enables generalization to unseen domains without fine-tuning.

\section{Ablation Study}

\begin{figure*}[ht!]
    \centering
    \includegraphics[width=0.7\linewidth]{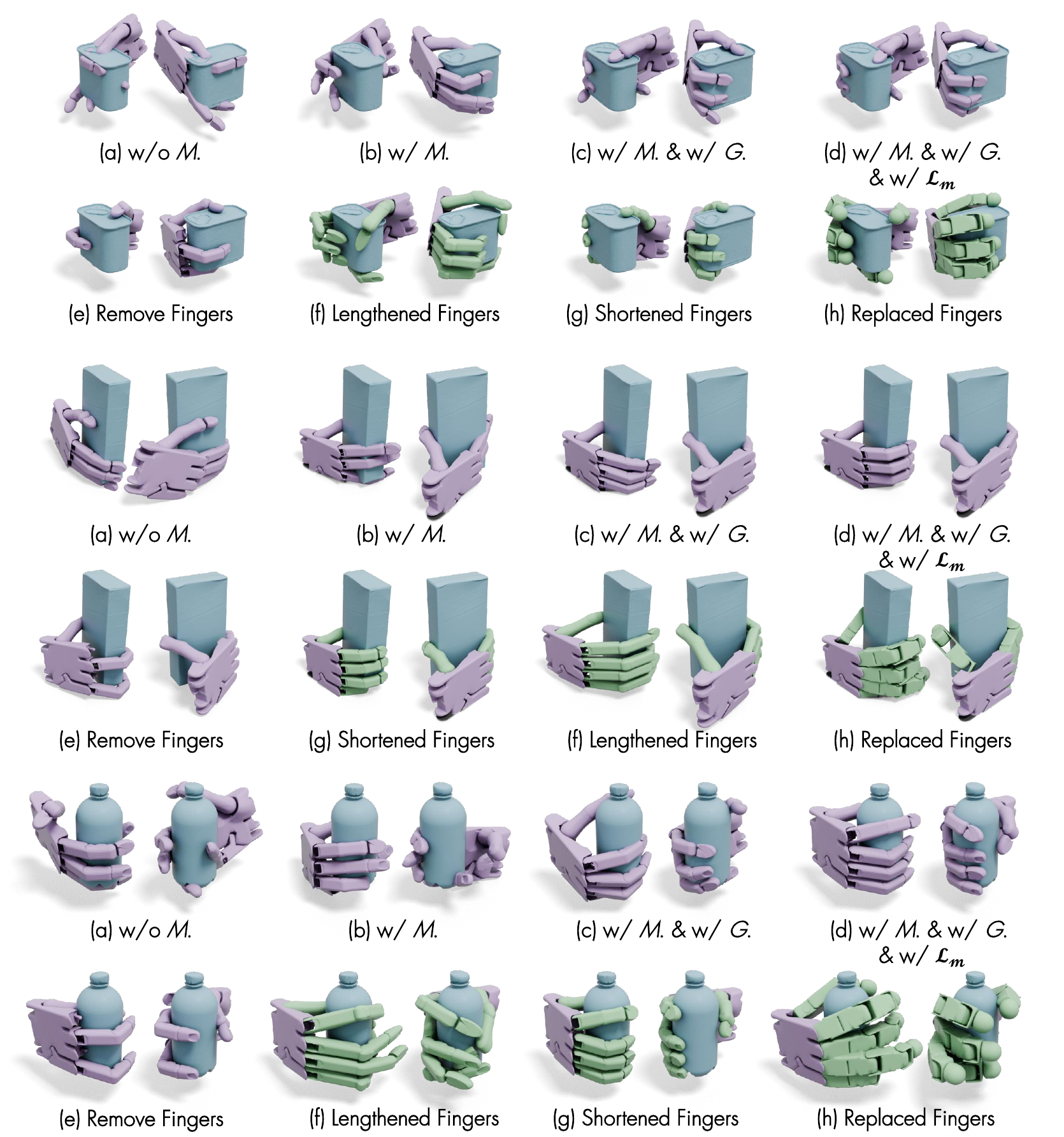}
    \vskip -0.6em
    \caption{ 
        Visualizations of ablation study on 1) effectiveness of morphology encoding and 2) zero-shot grasp generalization to novel hand morphologies based on the Shadow Hand. (a) w/o morphology encoding; (b) w/ morphology encoding; (c) w/ morphology encoding and graph attention; (d) w/ morphology encoding, graph attention, and morphological loss; (e)-(g) Altered fingers.
    }
    \label{fig:vis_ablation_all}
\end{figure*}

\noindent \textbf{Morphology Encoding} \label{supp:ablation_morph}
To validate the effectiveness of our proposed morphology encoding mechanism, we conduct ablation studies on the MultiDex \cite{li2023gendexgrasp} dataset of: 1) morph encoding in DiffMorphGrasp, and 2) the graph attention employed in the morphology encoder compared to cross-attention from a basic Transformer \cite{dosovitskiy2020vit}. Tab. \ref{tab:ablation_morph} demonstrates that incorporating morphology encoding results in a 7.3\% improvement in success rate and a 2.4\% increase in diversity. Employing the graph attention from hand kinematic trees, as opposed to cross-attention from a basic Transformer, further leads to an additional 1.8\% increase in success rate and a 1.0\% improvement in diversity. Adding morphology encoding increases inference time by only 0.02 seconds, which is negligible for real-time applications, while replacing the basic cross-attention with the graph attention does not affect efficiency. Qualitative results in Fig. \ref{fig:vis_ablation_all} use identical initial noisy poses for fair comparison, showing that incorporating morphology encoding significantly enhances grasp quality, generating more physically plausible and stable grasps. 

\begin{table*}[ht!]  
    \centering  
    \caption{
    Ablation results on morphology encoding on the MultiDex \cite{li2023gendexgrasp} dataset across robotic hands from three to five fingers: Barrett, Allegro, and Shadow hand. $M.$ refers to morph encoding, and $G.$ refers to the graph attention employed in the morphology encoder. }  
    \label{tab:ablation_morph}  
    \fontsize{7.5pt}{9pt}\selectfont
    \begin{tabular}{c c | c c c c | c c c c | c}
        \toprule  
        \multirow{2}{*}{$M.$} & \multirow{2}{*}{$G.$} & \multicolumn{4}{c}{Success Rate (\%) $\uparrow$} & \multicolumn{4}{c}{Diversity (rad.) $\uparrow$} & \multirow{2}{*}{Eff. (s) $\downarrow$}\\
        \cmidrule(lr){3-6} \cmidrule(lr){7-10}
        &  & Barrett & Allegro & ShadowHand & \textbf{Avg.} & Barrett & Allegro & ShadowHand & \textbf{Avg.} \\
        \midrule
        - & - & 83.7 & 78.8 & 87.5 & 83.3 & 0.669 & 0.431 & 0.430 & 0.510 & \textbf{0.45} \\
        \ding{51} & - & 90.6 & 86.9 & 94.4 & 90.6 & 0.678 & 0.438 & 0.450 & 0.522 & 0.47 \\
        \ding{51} & \ding{51} & \textbf{92.5} & \textbf{89.6} & \textbf{95.0} & \textbf{92.4} & \textbf{0.698} & \textbf{0.462} & \textbf{0.451} & \textbf{0.537} & 0.47 \\
        \bottomrule  
    \end{tabular}  
\end{table*}

\noindent \textbf{Adaptive Weight in Morphological Loss Function} \label{supp:geometric_mean}
We provide more details of adaptive weight $w_i$ in our morphological loss function. The design is rooted from robot kinematics. In a kinematic chain, angular errors at proximal joints propagate and compound down the chain, affecting all distal links. Therefore, joints with more descendants ($c_i$) require larger penalization. However, it can cause severe weight imbalance when using an arithmetic mean weighting $\frac{c_i+1}{S}$, where $S$ represents sum of joint and “+1” is used for numerical stability. For example, on the 24-DoF Shadow Hand, comparing the root joint $c_i=23$ with a proximal finger joint $c_i=23$, an arithmetic normalization creates a 6:1 weight ratio (24 vs 4). This penalizes the root while neglecting crucial finger articulations.

\begin{table*}[ht!]  
    \centering  
    \caption{
    Ablation results on adaptive weights in morphology loss on the MultiDex \cite{li2023gendexgrasp} dataset. We compare baseline (w/o. morphology loss), arithmetic mean, and geometric mean. }  
    \label{tab:geometric_mean}
    \fontsize{7.5pt}{9pt}\selectfont
    \begin{tabular}{l | c c c c | c c c c}
        \toprule  
         \multirow{2}{*}{Ablation} & \multicolumn{4}{c}{Success Rate (\%) $\uparrow$} & \multicolumn{4}{c}{Diversity (rad.) $\uparrow$} \\
        \cmidrule(lr){2-5} \cmidrule(lr){6-9}
        & Barrett & Allegro & ShadowHand & \textbf{Avg.} & Barrett & Allegro & ShadowHand & \textbf{Avg.} \\
        \midrule
        Baseline & 92.5 & 89.6 & 95.0 & 92.4 & 0.698 & 0.462 & \textbf{0.451} & \textbf{0.537} \\
        Arithmetic Mean & 91.9 & 88.8 & 91.2 & 90.6 & 0.683 & 0.443 & 0.439 & 0.522 \\
        Geometric Mean & \textbf{93.0} & \textbf{90.3} & \textbf{98.8} & \textbf{94.0} & \textbf{0.698} & \textbf{0.462} & 0.445 & 0.535 \\
        \bottomrule  
    \end{tabular}  
\end{table*}

In contrast, our proposed Eq. \ref{eq:adaptive_weight} introduces a square-root scaling mechanism. While both arithmetic and geometric normalizations can anchor the average weight at 1 to prevent overall loss scale drift, the arithmetic approach does so at the cost of extreme variance. Our non-linear scaling resolves this by reducing the harsh 6:1 disparity into a smoother $\sqrt{6}:1$ (2.45:1) ratio, ensuring distal joints are still sufficiently penalized. Furthermore, geometric mean $\mathcal{G}$ (Eq. \ref{eq:geometric_mean}) aligns with the multiplicative nature of kinematic error propagation down the chain. It guarantees that the geometric mean of the final applied weights remains 1, preserving optimization stability while elegantly resolving the weight imbalance. This is validated in Tab. \ref{tab:geometric_mean}. As expected, the arithmetic setting over-optimizes the root and neglects distal fingers, causing a significant performance drop, e.g., 98.8\% to 91.2\% for success rate on the Shadow hand. Conversely, the geometric mean setting balances the kinematic tree, achieving the best overall success rate (94.0\%) and diversity (0.535 rad).

\section{Zero-Shot Generalization to Structural Variations} \label{supp:more_generalize}

\begin{figure*}[ht!]
    \centering
    \includegraphics[width=\linewidth]{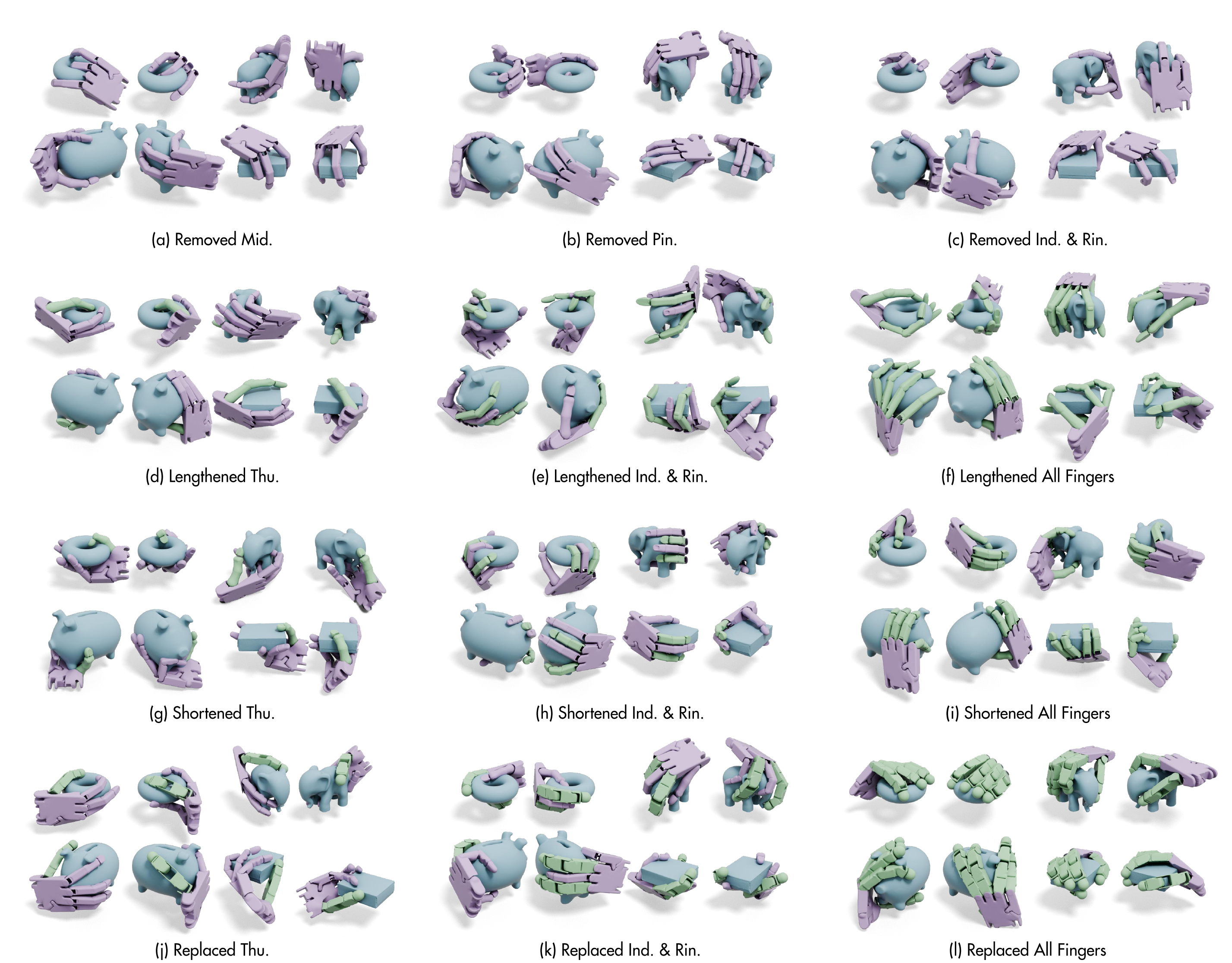}
    \vskip -0.6em
    \caption{ 
        More visualizations of zero-shot generalization to structural variations of the Shadow hand.
        \textbf{Topological variations:} (a) remove middle finger; (b) remove pinky; (c) remove index and ring. 
        \textbf{Geometrical variations:} (d) lengthen thumb; (e) lengthen index and ring; (f) lengthen all fingers; (g) shorten thumb; (h) shorten index and ring; (i) shorten all fingers. 
        \textbf{Embodiment variations:} (j) replace thumb with Allegro; (k) replace index and ring with Allegro; (l) replace all fingers with Allegro. Altered fingers are highlighted in green. 
    }
    \label{fig:vis_more_generalize}
\end{figure*}

We provide more details and results of zero-shot generalization to structural variations of the Shadow hand, including topological, geometrical, and hybrid embodiment variations.
Quantitative and qualitative results are shown in Tab. \ref{tab:ablation_fingers} and Fig. \ref{fig:vis_more_generalize}. 
Across all tested variations, the model maintains a consistently high success rate with performance degradation limited to 0.5\%-16.1\%, comprehensively validating its capability to generalize to novel hand structures without retraining. 

\begin{table}[ht!] 
    \centering  
    \caption{
    Ablation study on zero-shot generalization to structural variations of the Shadow hand using the MultiDex \cite{li2023gendexgrasp} dataset. $-$ indicates the finger remains the same. \textbf{Topological variations}: \ding{55} indicates the corresponding finger is removed; \textbf{Geometrical variations}: $>$ indicates the corresponding finger length is scaled by a factor of $1.5$, and $<$ indicates it is scaled by $0.8$; \textbf{Embodiment variations}: $\sim$ indicates the corresponding Shadow finger is replaced by an Allegro finger. 
    }  
    \label{tab:ablation_fingers}  
    \fontsize{7.5pt}{9pt}\selectfont
    \begin{tabular}{c c c c c c c}
        \toprule  
        \multirow{2}{*}{} & \multicolumn{5}{c}{Altered Fingers} & \multirow{2}{*}{Suc. Rat. (\%)}\\
        \cmidrule(lr){2-6}
        & Thu. & Ind. & Mid. & Rin. & Pin. & \\
        \midrule
        & - & - & - & - & - & \textbf{98.8} \\
        \hdashline
        \multirow{6}{*}{\rotatebox{90}{Topo.}}
        & - & \ding{55} & - & - & - & 93.2 \\
        & - & - & \ding{55} & - & - & 97.5 \\ 
        & - & - & - & \ding{55} & - & 98.3 \\
        & - & - & - & - & \ding{55} & 84.2 \\
        & - & \ding{55} & - & \ding{55} & - & 91.9 \\
        & - & - & \ding{55} & - & \ding{55} & 82.7 \\
        \hdashline
        \multirow{6}{*}{\rotatebox{90}{Geom.}}
        & $>$ & - & - & - & - & 96.3 \\
        & $<$ & - & - & - & - & 96.9 \\ 
        & - & $>$ & - & $>$ & - & 96.7 \\ 
        & - & $<$ & - & $<$ & - & 97.8 \\ 
        & $>$ & $>$ & $>$ & $>$ & $>$ & 92.5 \\ 
        & $<$ & $<$ & $<$ & $<$ & $<$ & 96.6 \\ 
        \hdashline
        \multirow{4}{*}{\rotatebox{90}{Embo.}}
        & - & - & - & - & - & 98.8  \\
        & $\sim$ & - & - & - & - & 97.9 \\
        & - & $\sim$ & - & $\sim$ & - & 90.6 \\ 
        & $\sim$ & $\sim$ & $\sim$ & $\sim$ & $\sim$ & 91.3 \\ 
        \bottomrule
    \end{tabular}  
\end{table}

\textbf{Topological Variations.} 
We first test our model's topological robustness by removing specific fingers. We retain the thumb to ensure force closure for opposable grasping. It can be observed that the removal of the pinky finger results in the most significant performance drop, whereas the absence of the middle or ring finger has a minimal impact. 

\textbf{Geometrical Variations.} 
Second, we test our model's geometrical robustness by altering the lengths of fingers, \textit{i.e.}, scaling the finger lengths by factors of $1.5\times$ (lengthened) and $0.8\times$ (shortened). 
Notably, lengthening the fingers leads to a more significant negative impact (up to 6.3\% drop) than shortening them. This is likely because excessive finger length introduces kinematic redundancy or potential self-collisions that complicate the formation of stable grasps.

\textbf{Embodiment Variations.} 
Finally, we evaluate the model's adaptability to cross-embodiment variations by creating hybrid hand structures, \textit{i.e.}, replacing Shadow Hand fingers with those from Allegro Hand to introduce embodiment changes in joint axis, joint limits, and link geometries. 
We observe that replacing the thumb has a negligible impact, whereas altering other fingers leads to a more distinct drop. This is likely due to the kinematic mismatch in hybrid embodiments, which introduces unexpected self-collisions or constraints that challenge grasp stability. 

We provide additional qualitative results in Fig. \ref{fig:vis_more_generalize} to validate our model's zero-shot generalization capability across diverse topological, geometrical, and embodiment variations of the Shadow Hand.

\section{Real-World Experiment}
\label{supp:real_world}

Our real-world setup uses a UR5e arm equipped with a Leap Hand or Allegro Hand. An Intel RealSense D435 captures RGB-D observations. We use SAM \cite{kirillov2023segment} to segment the object and extract a partial point cloud, followed by PCN \cite{yuan2018pcn} for point-cloud completion as input to DiffMorphGrasp. Each generated grasp is executed by moving the robot to the predicted wrist pose and commanding the hand to the predicted joint angles. A trial succeeds if the object is lifted without being dropped.

\begin{figure*}[ht!]
    \centering
    \includegraphics[width=\linewidth]{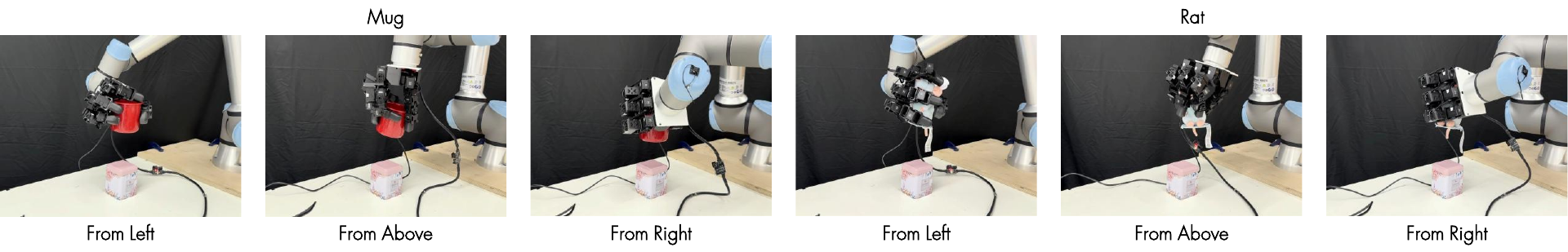}
    \vskip -0.6em
    \caption{
    Diverse grasping strategies from different approach directions in the real world.
    }
    \label{fig:direction}
\end{figure*}

For the YCB evaluation in Sec.~4.5, we conduct 10 trials per object using the Leap Hand. Fig.~\ref{fig:direction} further shows diverse grasping strategies from different approach directions.

\begin{figure*}[ht!]
    \centering
    \includegraphics[width=\linewidth]{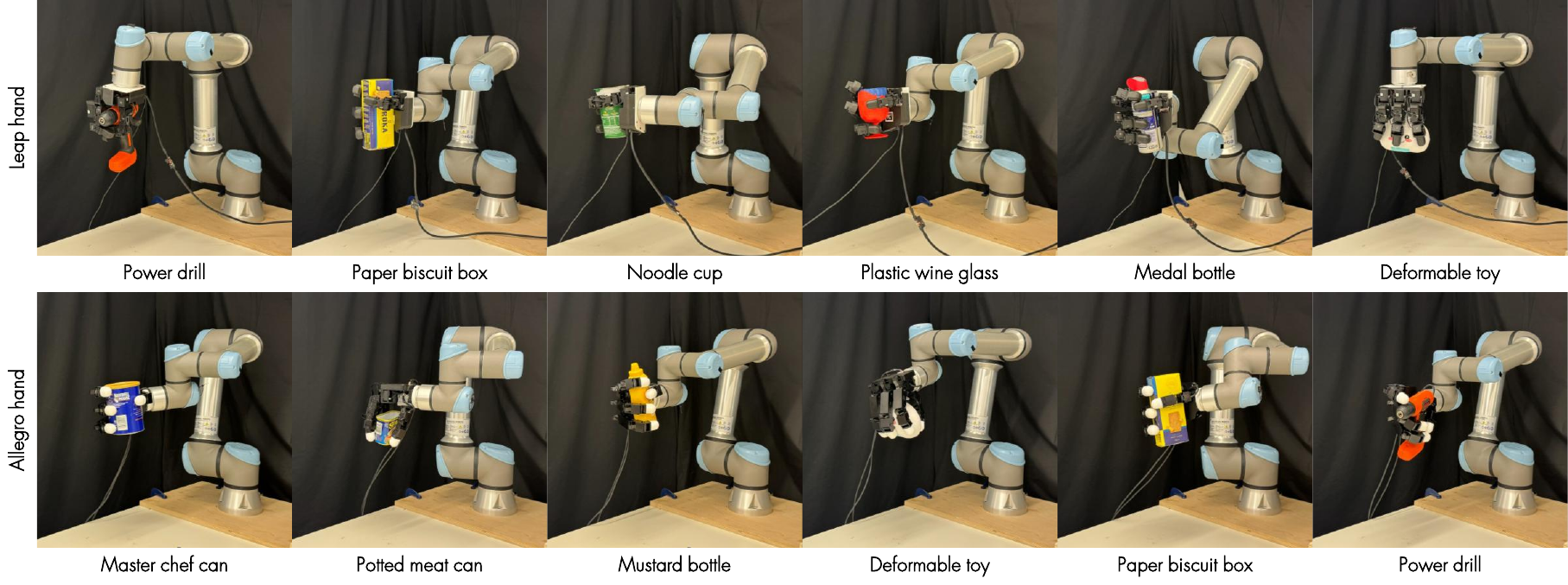}
    \vskip -0.6em
    \caption{
    Representative real-world grasps across Leap and Allegro hands.
    }
    \label{fig:cross_embodiment_real}
\end{figure*}

We further expand the evaluation to 14 objects, including six unseen everyday objects, using both Leap and Allegro Hands with 20 trials per object for each hand. Leap and Allegro achieve 86.8\% and 89.6\% overall success rates, respectively, with both hands successfully grasping all six unseen objects. Allegro performs better on large objects such as the power drill due to its larger hand size. Representative cross-embodiment grasps are shown in Fig.~\ref{fig:cross_embodiment_real}.

\section{Real-World Reactive Grasping in Dynamic Environments}
\label{supp:closed_loop}

\begin{figure*}[ht!]
    \centering
    \includegraphics[width=\linewidth]{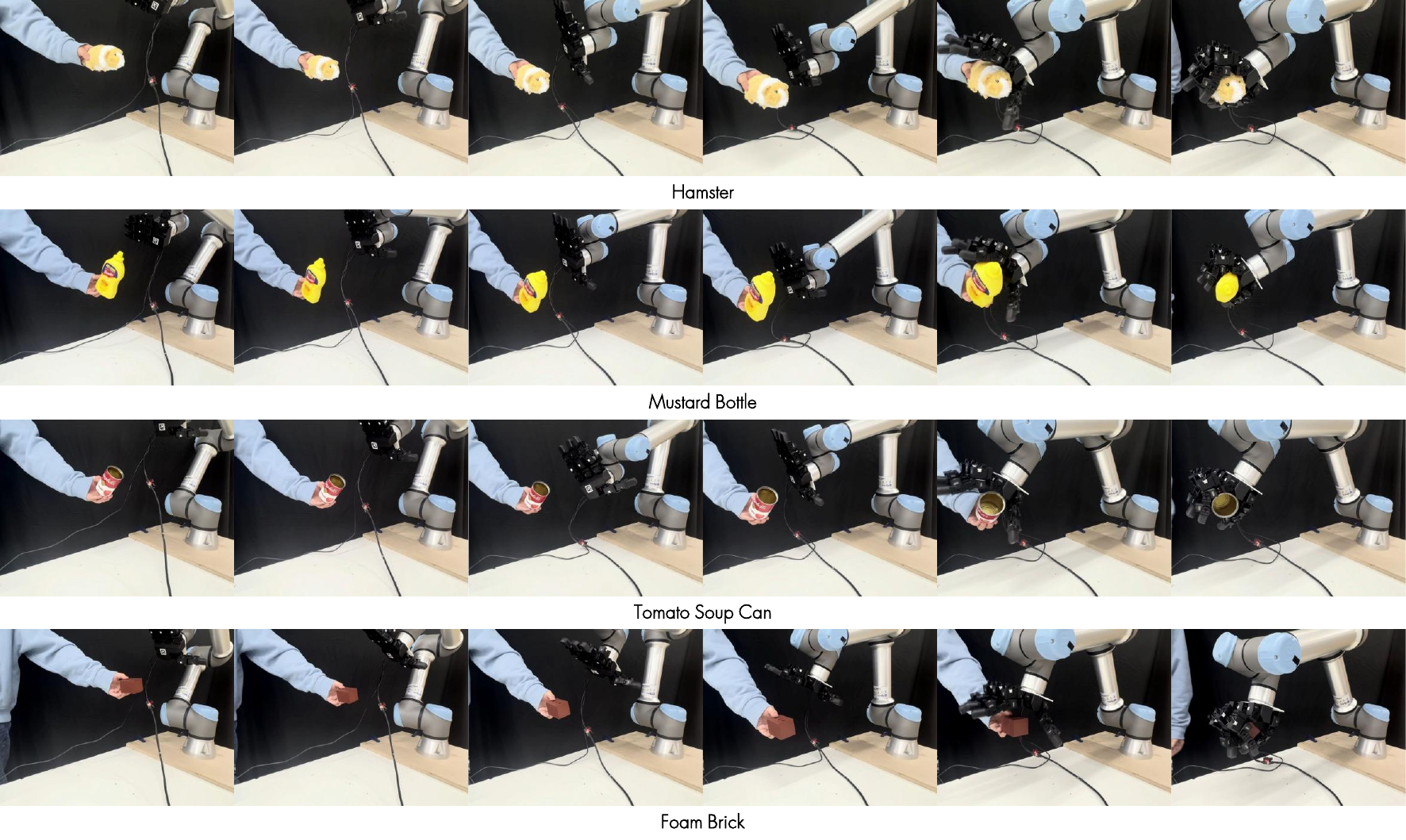}
    \vskip -0.6em
    \caption{ 
        Closed-loop reactive grasping in dynamic environments. In the scenario, a human holds and continuously moves the target object, and our framework generates updated grasp poses based on real-time point clouds captured by an RGB-D RealSense D435 camera. The robot's approaching motions from initial positions are also continuously guided by the generated poses.
    }
    \label{fig:closed_loop}
\end{figure*}

We demonstrate our model's capability of closed-loop reactive grasping in dynamic environments in Fig. \ref{fig:closed_loop} and demo video. In the experiment, a human holds the object and moves it in real-time, while an RGB-D RealSense D435 camera captures real-time point clouds to guide the grasp execution. 

\end{document}